%% file: neurips_2026.tex
\documentclass{article}

\usepackage[preprint]{neurips_2026}

\usepackage[utf8]{inputenc} % allow utf-8 input
\usepackage[T1]{fontenc}    % use 8-bit T1 fonts
\usepackage[pagebackref]{hyperref}       % hyperlinks
\usepackage{url}            % simple URL typesetting
\usepackage{booktabs}       % professional-quality tables
\usepackage{amsfonts}       % blackboard math symbols
\usepackage{nicefrac}       % compact symbols for 1/2, etc.
\usepackage{microtype}      % microtypography
\usepackage[table]{xcolor}
\usepackage[dvipsnames]{xcolor}

\usepackage{titletoc}
\usepackage{graphicx}
\usepackage{wrapfig}
\usepackage{enumitem}
\usepackage{multirow}
\usepackage{caption,subcaption}
\usepackage{amsmath,amssymb,amsthm}
\usepackage[capitalise]{cleveref}
\usepackage[rightcaption]{sidecap}
\usepackage{pifont}
\usepackage{fontawesome5}
\input{math_commands.tex}

\usepackage{algorithm}
\usepackage{algorithmic}

\crefformat{subsection}{\S#2#1#3}
\crefformat{section}{\S#2#1#3}
\newcommand{\cmark}{\textcolor{darkgreen}{\ding{51}}}%
\newcommand{\xmark}{\textcolor{darkred}{\ding{55}}}%
\definecolor{darkgreen}{rgb}{0.0,0.7,0.0}
\definecolor{darkred}{rgb}{0.7,0.0,0.0}
\hypersetup{
    colorlinks,
    linkcolor=RoyalBlue,
    citecolor=RoyalBlue,
    urlcolor=RoyalBlue,
    filecolor=RoyalBlue,
}
\renewcommand*{\backrefalt}[4]{%
    \ifcase #1 {(not cited)}%
    \or        {(cited on page~#2)}%
    \else      {(cited on pages~#2)}%
    \fi
}

\title{Beyond a \emph{Single Explanation} of the Adam--SGD Gap}

\author{%
  {\bfseries Chenxiang Zhang$^{1,3,}$\thanks{Co-first authors (see~\Cref{sec:contributions}). Correspondence: \texttt{chenxiang.zhang@uni.lu, antonio@tue.ellis.eu}}} \quad
  {\bfseries Rustem Islamov$^{2,*}$} \quad
  {\bfseries Enea Monzio Compagnoni$^{2}$} \\
  {\bfseries Jun Pang$^{1}$} \quad
  {\bfseries Aurelien Lucchi$^{2}$} \quad
  {\bfseries Antonio Orvieto$^{3,4,5}$}
  \vspace{0.25em}
  \\
  $^1$University of Luxembourg \quad $^2$University of Basel\\
  $^3$MPI for Intelligent Systems \quad $^4$ELLIS T\"{u}bingen \quad $^5$T\"{u}bingen AI Center
}

\begin{document}
\maketitle

\begin{abstract}
% Introduction & motivation
Prior work has identified several factors that can contribute to the performance gap between Adam and SGD, spanning data aspects, architecture design, and optimization properties. Yet these explanations are often studied in isolation, leaving their relative importance unclear. 
% Results
In this work, we revisit these hypotheses through a controlled empirical study across vision, language, genomics, and graph tasks, spanning modern and classical architectures, and carefully designed training setups. 
Our results suggest that no single factor consistently explains the Adam--SGD gap.
For instance, the Adam advantage can (1) persist under a uniform vocabulary distribution yet nearly disappear under a heavy-tailed one; (2) reverse in favor of SGD in softmax-attention models; and (3) become larger under soft architectural modifications, e.g., when ReLU is replaced by a GeLU nonlinearity.
% Impact
This suggests that the gap arises from nontrivial data and architecture interactions, rather than from a single common factor. Yet, we observe a pattern across our settings: a \emph{crossover batch size} at which the relative advantage shifts from SGD to Adam as the batch size scales. These empirical results are captured by our theoretical gap model, which predicts this batch-size-dependent crossover. Our perspective helps reconcile several existing hypotheses while offering practical insights across domains.  

\hfill \url{https://github.com/orientino/gap}
\end{abstract}

% ----------------------------------------------------
% ----------------------------------------------------
% ----------------------------------------------------

\vspace{-2em}
\section{Introduction}
\label{sec:intro}

\begin{figure}[!th]
  \begin{minipage}[c]{.31\linewidth}
    \vspace{-1em}
    \centering
    \scriptsize
    \begin{tabular}{llr}
      \toprule
      \textbf{Data} & \textbf{Vocab.} & \multicolumn{1}{l}{\textbf{Vocab.}} \\
      & \textbf{Distribution} & \multicolumn{1}{l}{\textbf{Size $v$}} \\
      \midrule
      FineWeb & Zipf & 50257 \\
      I21K & Zipf & 19167 \\
      HG38 & Uniform & 5 \\
      ZINC & - & - \\
      \bottomrule
    \end{tabular}
    \vspace{0.5em}
    \vfill
    \begin{tabular}{lll}
      \toprule
      \textbf{Data} & \textbf{Transf.} & \textbf{Non-Transf.} \\
                   & \textbf{Arch.}   & \textbf{Arch.} \\
      \midrule
      FineWeb & GPT  & GCNN  \\
      I21K    & ViT  & ResNet\\
      ZINC    & GRIT & GAT   \\
      HG38    & GPT  & GDN   \\
      \bottomrule
    \end{tabular}
  \end{minipage}%
  \begin{subfigure}[c]{.345\linewidth}
    \centering
    \includegraphics[width=\linewidth]{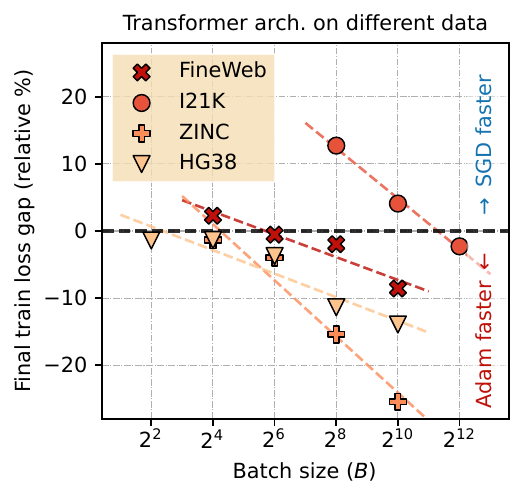}
  \end{subfigure}%
  \begin{subfigure}[c]{.33\linewidth}
    \centering
    \includegraphics[width=\linewidth]{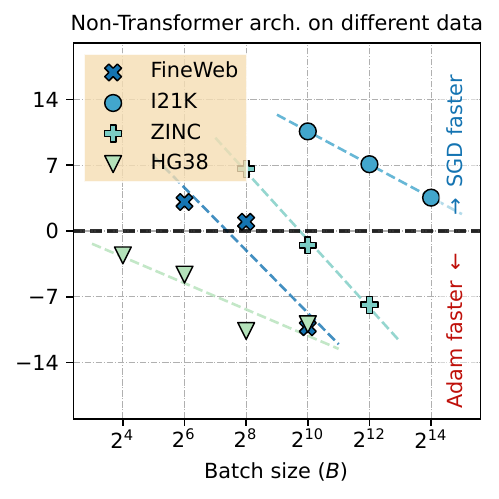}
  \end{subfigure}% 
  \caption{
\textbf{Optimizer advantage shifts from SGD to Adam as the batch size scales, with the trend depending on interactions between data and architecture.}
The $y$-axis reports the relative gap $(\mathcal{L}_{\text{Adam}} - \mathcal{L}_{\text{SGD}}) / \mathcal{L}_{\text{Adam}}$, where positive values indicate SGD outperforms Adam.
\textbf{(left)} Our experimental setup spans four data modalities with different vocabulary distributions and sizes, trained on both Transformer and non-Transformer architectures.
\textbf{(middle)} Transformer and \textbf{(right)} non-Transformer models trained across language, vision, genomics, and graph learning.
Across all settings, increasing batch size consistently shifts the advantage from SGD to Adam; however, the crossover point and transition slope vary by configuration. Rather than attributing the Adam--SGD gap to a single factor, these results point to a \emph{crossover batch size} whose value is jointly determined by data and architecture properties. In all the experiments, the learning rate and momentum are jointly and extensively tuned, see~\Cref{sec:method}.
  }
  \label{fig:main} 
  \vspace{-3mm}
\end{figure}

For over a decade, understanding which optimizer properties drive performance in deep learning has been a central question, particularly in Transformer-based language modeling (LLMs). At the scale of modern LLMs, optimization efficiency is crucial in terms of resource utilization. At present, most industrial and open-source models rely on adaptive methods such as Adam~\citep{kingma2014adam,loshchilov2018decoupled}, which have become the de facto standard. 
The success of this optimizer, however, is not limited to language or foundation models. Early benchmarks ~\citep{schmidt2021descending} report that Adam outperforms SGD (Adam$>$SGD) in simple VAEs trained on MNIST and F-MNIST. AlgoPerf~\citep{dahl2023benchmarking} also shows an advantage for Adam on tasks such as speech recognition with LSTMs and click-through-rate prediction with MLP-based recommender models. Recently, \citet{zucchet2024recurrent} show that large scale is also not a necessary ingredient, as simple recurrent networks on synthetic Gaussian data are faster to optimize with Adam than SGD.

% Current status
A growing body of work has made several advancements in explaining the Adam--SGD gap by isolating individual factors. While these studies provide valuable insights, their conclusions are often drawn within specific domains or training setups, making it difficult to assess their relative importance or understand how they interact across settings. To navigate this landscape, we group existing explanations into:
\begin{itemize}[leftmargin=*]

    \item \emph{Data source} explanations attribute the gap to data properties, in particular the Zipfian vocabulary distribution of text data, which induces a highly anisotropic optimization landscape~\citep{kunstner2024heavy,kunstner2025scaling}. In contrast, SGD has historically been successful on vision with CNNs~\citep{zhou2020towards,dahl2023benchmarking}, where labels are more uniformly distributed.
    
    \item \emph{Architecture source} explanations argue that the gap arises from the component heterogeneity of Transformers, including embedding~\citep{kumar2022finetune}, softmax-attention~\citep{noci2022signal,ahn2024linear}, normalization and head layers~\citep{zhao2025deconstructing}. 
    This perspective emphasizes that Adam's advantage is a consequence of the architectural complexity of modern networks. 

    \item \emph{Optimization source} explanations attribute the gap to batch size and gradient noise. In the small-batch regime, SGD can match or even outperform Adam in Transformer-based LLMs~\citep{sreckovic2025your,marek2025small}, suggesting that large-batch training in modern pipelines contributes to Adam's advantage. A separate line of work attributes Adam's advantage to its robustness under heavy-tailed gradient noise~\citep{cutkosky2021high,fatkhullin2025can,hubler2025gradient}.
    
\end{itemize}

These explanations are mostly motivated by observations from LLM training, each emphasizing a different source of the gap. In this paper, our primary goal is to connect these existing hypotheses. 

\subsection{Contributions}
\label{subsec:contributions}

We answer the following question: \emph{which of the existing factors genuinely drive the Adam--SGD gap, and do these explanations transfer outside the standard Transformer language modeling setting?} 

\textbf{We broaden the Adam--SGD gap study under a unified, modern, and controlled setup.} Our experiments cover classic domains such as language and vision, as well as underrepresented ones like genomics and graphs, with both Transformer and classical architectures. Importantly, we adopt the recent architectural advances and a modern optimization recipe while carefully controlling for confounders by jointly sweeping over the learning rate, momentum, and batch size~(\Cref{sec:method}).

\textbf{We find no single-factor hypothesis that fully explains the gap.} Vocabulary imbalance is not required: the gap favoring Adam persists on the genomics dataset HG38~\citep{hg38}, whose vocabulary is small and near-uniform~(\Cref{subsec:exp_data}). Softmax-attention is likewise not necessary~(\Cref{subsec:softmax}), and component heterogeneity is not sufficient: a Vision Transformer trained on ImageNet21K can exhibit the reverse trend, with SGD outperforming Adam. Rather than a sharp Transformer versus non-Transformer effect, the architectural contribution shifts gradually with individual design choices; e.g., replacing ReLU with GeLU already moves the advantage toward Adam~(\Cref{subsec:exp_arch}).

\textbf{A \emph{crossover batch size} connects the existing single-factor explanations.} Building on recent observations that the Adam--SGD gap in language modeling has a batch size dependence~\citep{sreckovic2025your,marek2025small}, we show that, under a fixed number of training samples, the advantage shifts from SGD to Adam as the batch size scales~(\Cref{fig:main}). Both data properties and architectural design interact with this phase change -- including nuances such as the nonlinearity function. Notably, this transition occurs at substantially larger batch sizes for image data regardless of the architecture, consistent with the strong empirical performance of SGD on vision tasks.

\textbf{A theoretical gap model captures this batch-size dependence.} We complement the empirical study with a simple geometry-based model of how the gap changes with batch size. Since all experiments use global gradient-norm clipping, we compare normalized SGD with momentum to a signed-momentum proxy for Adam, motivated by the connection between Adam and SignSGD~\citep{balles2018dissecting,orvieto2025search}. This places both methods in the non-Euclidean stochastic-update framework of~\citet{kovalev2025understanding}. Depending on the ratio of the smoothness constants defined with respect to different norms and the gradient-noise variance, different scenarios are predicted: either a crossover from SGD to Adam occurs as the batch size increases, or Adam or SGD can dominate for all batch sizes~(\Cref{sec:theory}). This model is consistent with the numerical results.

\textbf{The gap behaves qualitatively differently across training regimes.} In the classical deep learning setting of overparametrized multi-epoch training, where the optimization problem is solvable, the gap shrinks as training can approach zero loss; in the modern sub-one-epoch underparametrized regime of LLMs, the gap persists throughout training~(\Cref{subsec:exp_opt}).

% ----------------------------------------------------

\section{Preliminaries / Related work}
\label{sec:related}

We introduce and categorize existing Adam--SGD gap literature under four groups.

\textbf{Optimization explanations: gradient noise imbalance and large batch size.}
\label{subsec:opt_source}
Classical literature shows a nontrivial interaction between gradient noise and preconditioning~\citep{kohler2017sub}.~\citet{zhang2019algorithmic} studied how preconditioning alters the critical batch size, hinting at properties that are related to our study. Yet, Adam cannot be put in direct relation to classical preconditioning~\citep{kunstner2019limitations,balles2018dissecting}. Recent works take a different approach, claiming that Adam’s advantage comes from its coordinate-wise scaling, which better handles gradient stochasticity than SGD. For example,~\citet{zhang2020adaptive} observed that Adam outperforms SGD when the stochastic gradient norm follows a heavy-tailed distribution for Transformers. Adam, being closely related to SignSGD~\citep{bernstein2018signsgd}, can naturally clip noise and stabilize updates. Complementarily,~\citet{fatkhullin2025can} demonstrate that SGD is provably slower under heavy-tailed noise than adaptive methods, and that normalization can achieve optimal theoretical convergence~\citep{cutkosky2021high,hubler2025gradient}. However,~\citet{kunstner2023noise} challenge this explanation, showing that Adam has the advantage even in deterministic full-batch settings. Recently,~\citet{sreckovic2025your,marek2025small} provide additional evidence: SGD can match, or even outperform, Adam at small batch sizes for Transformers. In support,~\citet{compagnoni2025adaptive} provide theoretical evidence that the convergence rate of SignSGD improves with increasing batch size, whereas SGD does not. 
Our findings extend these works. Across a wide range of setups, we show that a small batch size reduces Adam's advantage over SGD, and can even flip the advantage depending on the data and architecture properties. 

\textbf{Data explanations: heavy-tailed class imbalance and vocabulary size.}
\label{subsec:data_source}
Moving one layer above the noise hypothesis,~\citet{kunstner2024heavy} present heavy-tailed class imbalance as the source of the Adam--SGD gap. 
They showed that SGD reduces the loss of frequent classes faster than that of rare classes, also visible as slower average loss convergence. 
Therefore, the advantage of Adam and SignSGD-like methods may be attributed to a lower sensitivity to class imbalance.~\citet{kunstner2025scaling} formalize this idea using a linear bigram model with Zipf-distributed data, proving that the gain of adaptive methods over full-batch GD scales with vocabulary size $v$.
While compelling, our results suggest that \emph{vocabulary-based hypotheses are not necessary for observing the Adam--SGD gap.} A nontrivial gap exists for genomics data with near-uniform token distribution and a small vocabulary size ($v=5$). This setup represents the opposite scenario proposed by the class imbalance hypothesis, yet the gap still exists.~\citet{schaipp2025optimization} recently reported the existence of the Adam--SGD gap in diffusion model training, where no clear notion of class imbalance exists. Similarly, we also show that the gap persists for regression tasks, which have no class structure. 

\textbf{Architecture explanations: Hessian heterogeneity of Transformers.}
\label{subsec:arch_source}
A parallel line of work attributes the advantage of adaptive optimizers to architectural properties.~\citet{zhang2024transformers} argue that Transformers induce stronger Hessian heterogeneity across layers than CNNs. For the attention layers,~\citet{noci2022signal} connect the gap to the different training dynamics between Q, K, and V layers. In language modeling,~\citet{zhao2025deconstructing} emphasize the necessity of tuning the head and normalization layers with Adam. On the other hand, in transfer-learning vision tasks,~\citet{kumar2022finetune} demonstrate that freezing the embedding layer is sufficient to eliminate Adam's advantage. Even the role of the softmax is disputed:~\citet{ahn2024linear} show that attention without softmax can still reproduce the Adam--SGD gap.
The presented literature consistently indicates that architecture matters, but points to different mechanisms. We aim to clarify which ones are necessary. First, we replace the attention layer of GPT with gated convolution~\citep{dauphin2017language} and observe that the gap persists,  suggesting that softmax-attention may not be necessary. 
Then, we show that architecture design alone can continuously shift the advantage from SGD to Adam, with the choice of GeLU and LayerNorm favoring Adam~(\Cref{tab:arch_interpolation}), highlighting that the effect of architecture on the gap interacts with the data properties. 

\textbf{Insights and limitations of optimization benchmarks.}
Optimization benchmarks~\citep{schmidt2021descending,dahl2023benchmarking,kasimbeg2025accelerating} provide valuable empirical signals, but their conclusions depend on metrics, tuning budgets, schedules, batch-size policies, and hardware.
The resulting evidence is mixed: after tuning, Adam-family methods are often strong in machine translation, generative modeling, speech recognition, and recommender systems; SGD-family methods can instead win in classical convolutional image classification, graph prediction, and MRI reconstruction.
Notably, batch size is treated differently across benchmarks:~\citet{schmidt2021descending} fix it, while AlgoPerf~\citep{dahl2023benchmarking} treats it as task-specific and tunable as part of the submitted training algorithm, where it interacts with wall-clock performance on the target hardware. 
As a result, neither isolates batch size as a potential confounder. Here, we instead treat batch size as a central hyperparameter and study how it affects the sign and magnitude of the gap.
% While modern hardware makes large batches increasingly appealing, understanding these trends remains crucial for improving current strategies.

% ----------------------------------------------------

\section{Methodology}
\label{sec:method}

We define the Adam--SGD gap and describe the common experimental details used across the paper. 

\textbf{Problem setup.}
The main objective is to study the optimization gap between Adam and SGD at the end of training, while controlling the confounders listed in~\Cref{sec:related}. For each configuration, we sweep the learning rate $\eta$ and momentum $\beta$. When referring to SGD, we always mean SGD with momentum. 
Throughout this work, the focus is on minimizing the training loss rather than improving generalization (although results are reported in~\Cref{app:exp_validation}). Unless specified otherwise, we apply the EMA to smooth the noisy training loss history and report the last value, which is particularly useful for small batch size training. 
The gap $\Delta^* = \mathcal{L}_{\text{Adam}}^* - \mathcal{L}_{\text{SGD}}^*$ measures the final loss difference between the best runs across all the $\eta$ and $\beta$ sweeps, where 
\begin{center}
    \definecolor{wheat}{RGB}{245,222,179}
    \colorbox{wheat!50}{$\Delta^* > 0$ indicates SGD achieves lower final loss than Adam, and $\Delta^* < 0$ indicates otherwise.}
\end{center}

\textbf{Architecture setup.}
We systematically analyze Transformer-based architectures applied to language, vision, graphs, and genomics data domains. To control for architecture source~(\Cref{subsec:arch_source}), we also study classic attention-free architectures. Appendix~\Cref{tab:arch} presents an overview of all the architectures and datasets. 
Unless specified otherwise, for GPT and ViT, we adopt the recent architectural advances such as RoPE~\citep{su2024roformer}, RMSNorm~\citep{zhang2019root}, QKNorm~\citep{touvron2023llama}, EmbedNorm~\citep{loshchilov2025ngpt}, and the ReLU$^{2}$ nonlinearity~\citep{so2021searching,zhang2024relu}. For GCNN, we modify the GPT architecture by replacing the attention layers with the gated convolution layer~\citep{dauphin2017language}. The remaining architectures use their original implementations. A complete description of the experiments is in~\Cref{app:exp_setup}.

\textbf{Optimization tuning.} The recipe is common across all the setups, unless modifications are specifically reported. We use \texttt{bfloat16} precision, 10\% linear warmup with a cosine scheduler to 0, and global gradient norm clipping to 1 for Adam and SGD. Weight decay is not applied in the main experiments as it introduces another confounder, while blowing up the hyperparameter search space. Yet, for completeness, a subset of the results in \Cref{fig:main} is repeated with weight decay in~\Cref{app:exp_wd}, with consistent findings. 
For SGD, we jointly tune both the learning rate and the momentum for each batch size and training configuration. 
For Adam, jointly tuning $(\eta,\beta_1,\beta_2)$ would lead to a much larger search space, while fixing $(\beta_1,\beta_2)$ to standard defaults such as $(0.9,0.95)$~\citep{biderman2023pythia} can be suboptimal~\citep{orvieto2025search,zhang_how_2025}. Recent work further shows that the optimal momentum parameters depend on the training budget and batch size~\citep{shulgin2026deriving,marek2025small}: \citet{marek2025small} find that the best $\beta_2$ increases as batch size decreases, and their Figure~4 suggests a related trend for $\beta_1$. This batch-size dependence is also predicted by theoretical analyses~\citep{malladi2022sdes,compagnoni2025adaptive}, suggesting that both $\beta_1$ and $\beta_2$ should be adapted when the batch size changes. We therefore tune Adam's momentum, but use the tied parameterization $\beta_1=\beta_2$~(\textit{which we denote as }$\beta$) following \citet{orvieto2025search}. This keeps a comparable tuning budget to SGD in terms of learning rate and one momentum parameter, while connecting Adam to SignSGD with momentum and to the non-Euclidean analysis in \Cref{sec:theory}. We note, however, that for batch sizes approaching $1$, independently tuning $(\beta_1,\beta_2)$ can further boost Adam performance, bringing it closer to SGD~(cf.~\Cref{fig:main}); we report this ablation in \Cref{app:independent_betas}. This setup goes beyond the tied-momentum parameterization used in our theoretical analysis, and is less directly aligned with one-momentum methods such as SignSGD with momentum and Muon~\citep{jordan2024muon}, for which the single momentum parameter naturally corresponds to the constraint $\beta_1=\beta_2$~\citep{orvieto2025search}.

% ----------------------------------------------------

\section{Experiments}
\label{sec:exp}

In this section, we present experimental results revisiting existing gap explanations. This includes the role of softmax-attention, vocabulary distribution, architecture heterogeneity, and batch size.

\subsection{Softmax-attention hypothesis.}
\label{subsec:softmax}
We investigate whether the gap primarily arises from the coupled $Q,K,V$ interactions and normalization induced by softmax-attention~\citep{noci2022signal,ahn2024linear}. To this end, we replace all the GPT's softmax-attention layers with gated-convolution token mixers~(GCNN, \citet{dauphin2017language}), a competitive language modeling approach before the advent of Transformers. Our GCNN architecture, interleaved with pure feedforward layers, mixes tokens only using stacked causal 1D convolutions, as follows: $h(X) = (X * W + b) \otimes \sigma(X * V + c)$.
\begin{figure}[h]
  \centering
  \begin{minipage}[c]{0.55\linewidth}
  \begin{subfigure}[c]{.40\linewidth}
    \centering
    \includegraphics[width=\linewidth]{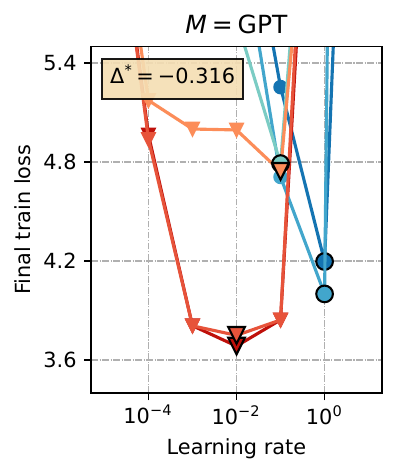}
  \end{subfigure}%
  \begin{subfigure}[c]{.60\linewidth}
    \centering
    \includegraphics[width=\linewidth]{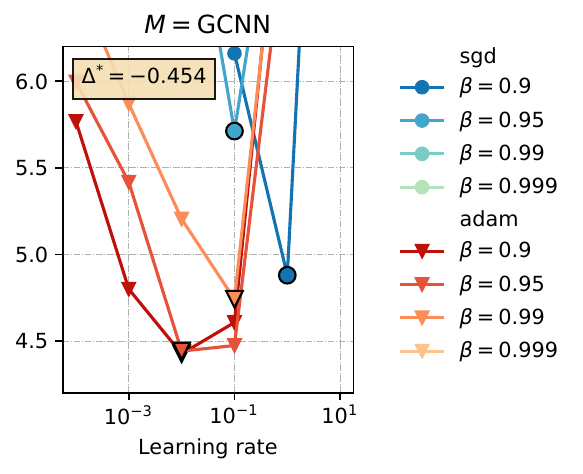}
  \end{subfigure}
  \end{minipage}
  \hfill
  \begin{minipage}[c]{0.44\linewidth}
    \caption{GPT and GCNN of 50M parameters trained on 1B FineWeb tokens at $B=1024$. GCNN only replaces the softmax-attention of GPT with a gated convolution along the sequence length. The Adam--SGD gap persists in both architectures, suggesting that softmax-attention is not necessary for the gap. $\Delta^*$ denotes the gap between the best Adam and SGD runs. Both the learning rate and momentum are tuned, see~\Cref{app:exp_gcnn}.}
    \label{fig:gap_gcnn}
  \end{minipage}
\end{figure}

% Result
\Cref{fig:gap_gcnn} shows that GCNN trained on FineWeb still has a wide gap $\Delta^*=-0.454$ despite the removal of the softmax-attention. 
SGD achieves $\approx8\%$ higher final loss than Adam for GPT, and $\approx10\%$ higher for GCNN. 
This experiment probes the role of softmax-attention: the persistence of the gap in GCNN suggests that Adam's advantage is neither specific to Transformers nor to the normalization induced by softmax, but extends to simpler convolution-based sequence models on text. Same conclusion is observed for state-space models with linear attention, see Appendix~\Cref{fig:fineweb_gdn_train_loss}. These results complement ``Observation 2'' by~\citet{zhang2024transformers}, who show that Adam can outperform SGD on MLP-Mixer architectures for vision tasks, where the Hessian exhibits block heterogeneity. Our findings reinforce this claim in a standardized LM setup through optimizer tuning informed by contemporary research\footnote{\citet{zhang2024transformers} report default PyTorch settings: $(\beta_1,\beta_2)=(0.9, 0.999)$ for Adam and $\beta=0.9$ for SGD. This choice may be suboptimal: ~\citet{zhao2025deconstructing,sreckovic2025your} suggest a much higher momentum for SGD.}.
The persistence of the gap without softmax-attention shifts the question from whether softmax-attention is necessary to which architectural components modulate the gap. Indeed, block-level heterogeneity in both architectures can be caused by several other components, which we study later in~\Cref{subsec:exp_arch}.

\subsection{Heavy-tailed vocabulary imbalance and size hypothesis}
\label{subsec:exp_data}
In~\Cref{subsec:softmax}, we found that the Adam--SGD gap persists across distinct token-mixing strategies. Here, we investigate whether the gap arises from the nature of language data itself. We show that neither a heavy-tailed vocabulary distribution nor a large vocabulary is necessary for the gap to appear.

\begin{figure}[!ht]
  \centering
  \hspace*{-2cm}
  \begin{subfigure}[c]{.27\linewidth}
    \centering
    \includegraphics[width=\linewidth]{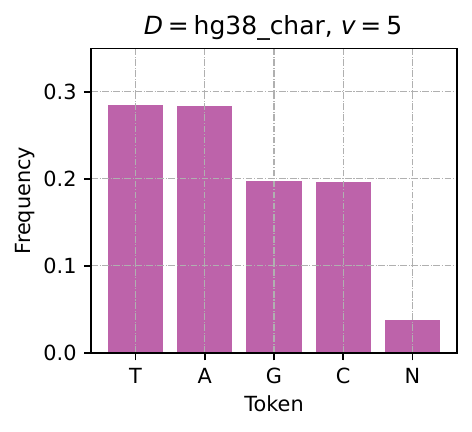}
  \end{subfigure}%
  \begin{subfigure}[c]{.27\linewidth}
    \centering
    \includegraphics[width=\linewidth]{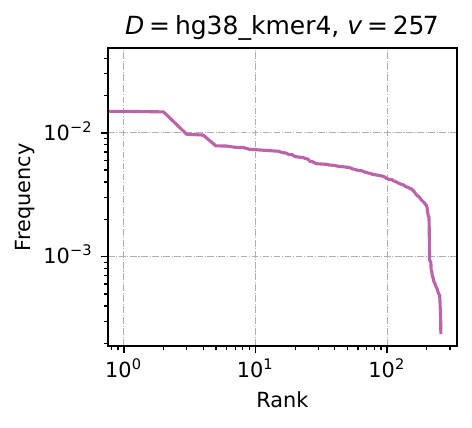}
  \end{subfigure}
  \begin{subfigure}[c]{.27\linewidth}
    \centering
    \includegraphics[width=\linewidth]{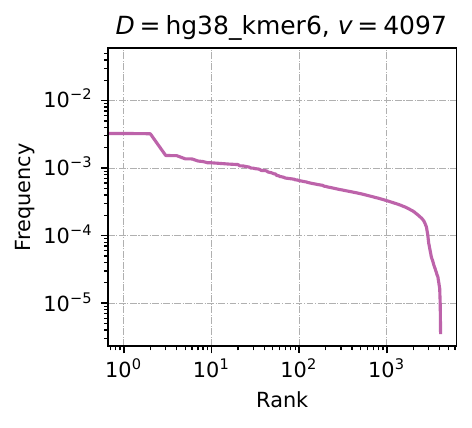}
  \end{subfigure}
  \begin{subfigure}[c]{.27\linewidth}
    \centering
    \includegraphics[width=\linewidth]{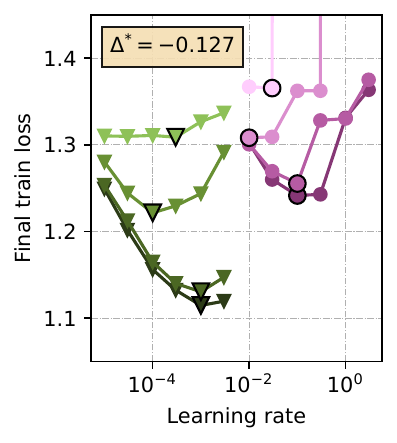}
  \end{subfigure}%
  \begin{subfigure}[c]{.27\linewidth}
    \centering
    \includegraphics[width=\linewidth]{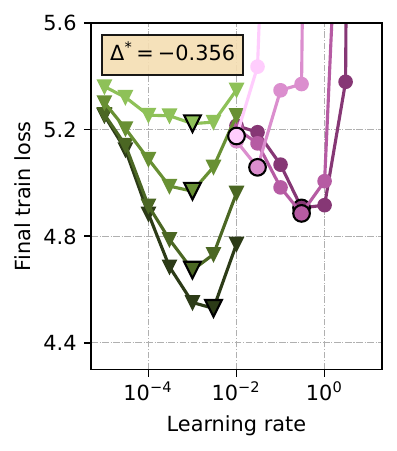}
  \end{subfigure}
  \begin{subfigure}[c]{.42\linewidth}
    \centering
    \includegraphics[width=\linewidth]{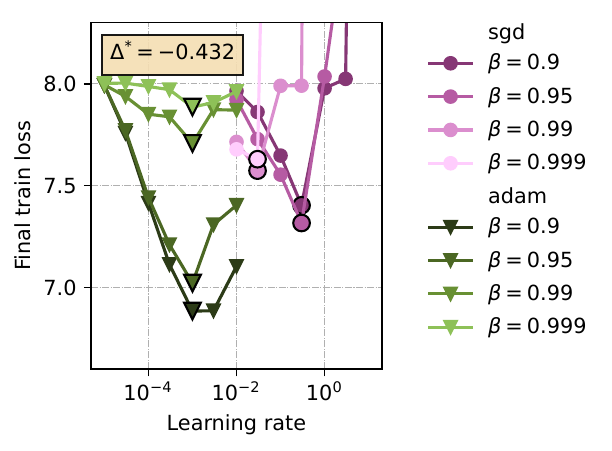}
  \end{subfigure}
  \caption{GPT trained on human genomics HG38 data processed with different tokenizers at batch size $B=256$. \emph{The Adam--SGD gap is present regardless of the vocabulary distribution and size.} In particular, under the \texttt{char} tokenizer with 2.7B tokens and a near-uniformly distributed small vocabulary ($v=5$), the gap is nontrivial. When the \texttt{kmer} tokenizer is applied, the number of tokens decreases, the vocabulary size grows, and its distribution becomes Zipfian with the Adam--SGD gap persisting. For every configuration, the learning rate and momentum are tuned, see~\Cref{app:exp_hg38}.}
  \label{fig:gap_hg38}
\end{figure}

\begin{wrapfigure}[20]{r}{0.3\linewidth}
  \centering
  \vspace{-5mm}
  \includegraphics[width=0.9\linewidth]{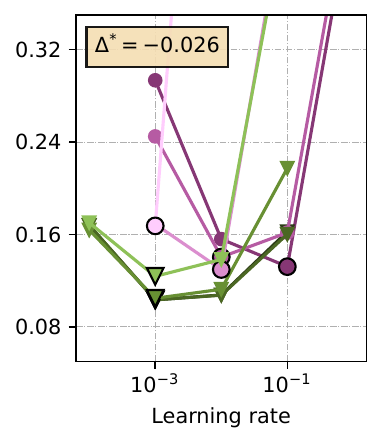}
  \caption{Graph Transformer trained on ZINC250K, a molecular regression task, at batch size $1024$. Adam>SGD despite this being a regression task with no class labels. The learning rate and momentum are tuned, see~\Cref{app:exp_hg38}.}
  \label{fig:gap_zinc250k_grit}
\end{wrapfigure}

\textbf{Vocabulary imbalance and size.}
We revisit the role of the heavy-tailed vocabulary~\citep{kunstner2024heavy} and the vocabulary size~\citep{kunstner2025scaling}. We train a GPT model on human genomics data HG38~\citep{hg38,kelley2020cross}, which enables fine control of the vocabulary distribution and size $v$ by changing the tokenizer between \texttt{char} and \texttt{kmer} tokenizer (i.e., single or non-overlapping groups of $k$ nucleotides). 
% This leads to three different scenarios:
% \begin{enumerate}[leftmargin=*, label=\alph*)]
%     \item \texttt{char} tokenizer: 10.63M parameter model, 2.72B tokens, $\approx$ uniform class distribution.
%     \item \texttt{4mer} tokenizer: 10.82M parameter model, 0.68B tokens, $\approx$ Zipfian class distribution. 
%     \item \texttt{6mer} tokenizer: 13.77M parameter model, 0.45B tokens, $\approx$ Zipfian class distribution.
% \end{enumerate}

% Results
\Cref{fig:gap_hg38} shows that, regardless of the vocabulary distribution and size $v$, the gap persists. Even with a \texttt{char}-level tokenizer, where the distribution is almost uniform (excluding the unsequenced nucleotides `N') with a small vocabulary $v=5$, the gap is in favor of Adam $\Delta^*=-0.127$, which translates into $\approx11\%$ SGD slowdown. For a larger vocabulary $v=257$ or $v=4097$, the gap remains as expected. 
These results show that vocabulary imbalance and large vocabulary size alone do not fully explain the Adam--SGD gap. Next, we present further evidence along the same line.

\textbf{Vocabulary in regression tasks.} We show that, even in regression tasks where \textit{no class structure is defined}, the gap can still exist. We train GRIT~\citep{ma2023grit}, a graph-Transformer model, on the real-world molecular dataset ZINC250K~\citep{irwin2012zinc}, where the task is to regress the constrained solubility with heavy atoms as nodes and atomic bonds as edges. The metric used here is MAE (Mean Absolute Error). We follow the graph learning literature and train for 100 epochs while sweeping both the learning rate and momentum, see~\Cref{app:exp_zinc}. 
% Result
\Cref{fig:gap_zinc250k_grit} shows that the absolute gap between the best SGD and Adam runs is $\Delta=-0.026$, which means that the best run of SGD achieves $\approx20\%$ worse final loss than Adam. This result complements~\citet{schaipp2025optimization}, who reports a similar gap in diffusion tasks, another setting without class structure. 
% Next, we move on to a imbalanced vision setting.

\begin{figure}[!t]
  \centering
  \begin{subfigure}[c]{.44\linewidth}
    \centering
    \includegraphics[width=0.50\linewidth]{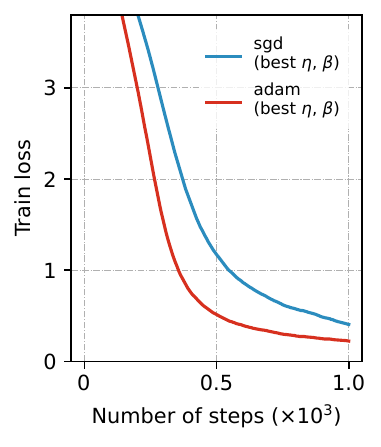}
    \includegraphics[width=0.46\linewidth]{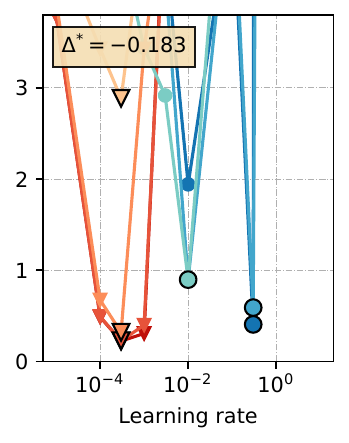}
    \caption{SimpleViT~(\citet{kunstner2024heavy})}
    \label{fig:gap_i1k_imb_simplevit}
  \end{subfigure}%
  \begin{subfigure}[c]{.56\linewidth}
    \centering
    \includegraphics[width=.39\linewidth]{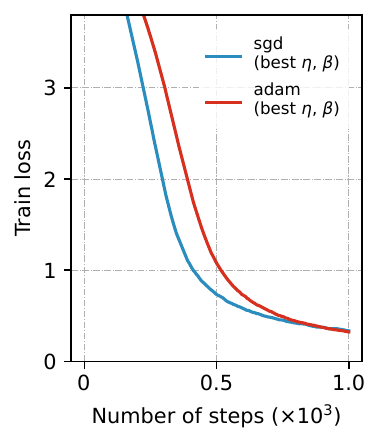}
    \includegraphics[width=.58\linewidth]{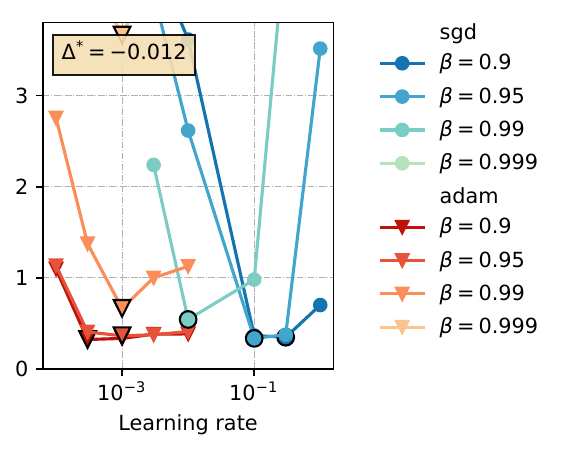}
    \caption{ViT + gradient norm clipping (ours)}
    \label{fig:gap_i1k_imb_vit}
  \end{subfigure}%
  \caption{ViT trained on Heavy-Tailed ImageNet1K at $B=1024$. \textbf{(a)} Reproduction of~\citet{kunstner2024heavy} shows a clear gap. \textbf{(b)} Modernizing the ViT and adding gradient norm clipping reduces the gap by $\approx93\%$. Notably, in the training curve of \textbf{(b)} SGD stays below Adam for most of training, the opposite of \textbf{(a)}, highlighting the effect of architecture and optimization recipe. Learning rate and momentum $\beta_1$ are tuned, see~\Cref{app:exp_ht_i1k}.}
  \label{fig:gap_i1k_imb}
\end{figure}

\textbf{Vocabulary in vision tasks.}
Following the same vision setup as~\citet{kunstner2024heavy}, we use the heavy-tailed ImageNet1K (HT-I1K) dataset, a subset of I1K with a heavy-tailed class distribution. 
We train a 12-layer ViT for 300 epochs with constant learning rate, $B=1024$, fixed $\beta_2=0.999$, and gradient norm clipping to 1 (more details in~\Cref{app:exp_ht_i1k}).~\Cref{fig:gap_i1k_imb_vit} shows that the gap is $\Delta=-0.012$, indicating that SGD is only $\approx3\%$ slower than Adam. Interestingly, the training curve shows that SGD achieves lower loss than Adam throughout most of training, falling behind only towards the end.
Note that our result here is in apparent contrast to~\citet{kunstner2024heavy}, who show that class imbalance causes Adam to have a clear advantage over SGD in this setup. This difference may reflect the use of gradient norm clipping plus different architectural considerations: we adopt recent design choices (e.g., ReLU$^{2}$ instead of GeLU, see~\Cref{app:arch}) and a convolutional \texttt{stem} instead of a linear layer, which helps to stabilize the training dynamics~\citep{xiao2021early}. These changes together bring the original gap from $\Delta^*=-0.183$ to $\Delta^*=-0.012$, a reduction of $\approx93\%$ in training loss gap. 
The cumulative results presented in this section indicate that class imbalance alone does not consistently produce the gap; rather, its effect is coupled with architecture and training choices. This motivates us to examine architectural components in the next section.

\subsection{Dissecting architectural components}
\label{subsec:exp_arch}

Beyond softmax-attention results in~\Cref{subsec:softmax}, recent works attribute the Adam--SGD gap to Transformer component heterogeneity.~\citet{zhang2024transformers} link it to block-level Hessian heterogeneity, while~\citet{zhao2025deconstructing} localize Adam's advantage to the head and normalization layers.

\input{tables/arch_hybrid}
\textbf{Embedding, normalization, and head layers.}
To revisit whether additional architectural heterogeneity contributes to the Adam--SGD gap, we broaden the set of experiments proposed by~\citet{zhao2025deconstructing}. Similarly, we train specific layers such as the head, embedding, normalization, or their combinations with Adam, while optimizing all remaining parameters with SGD. The reported loss is averaged over the last 100 iterations. For each configuration, we \emph{jointly} tune the learning rate $\eta$ and momentum $\beta$, see~\Cref{app:exp_hybrid}.
\Cref{tab:hybrid_optimizer} reveals a clear hierarchy among components. Applying Adam only to the normalization layers yields little speedup relative to using SGD for all layers. In contrast, using Adam for the embedding layer reduces the gap by $33\%$, extending the vision results from~\citet{kumar2022finetune}. Applying Adam to the output head yields an even larger improvement, reducing the gap by $69\%$, as in~\citep{zhao2025deconstructing}. Combining Adam on several of these layers can narrow the gap further, but does not eliminate it.
These findings suggest that modern architecture design contributes to the Adam--SGD gap. Yet, component heterogeneity alone does not necessarily imply Adam$>$SGD, as we illustrate with an ablation in the vision domain. ~\Cref{fig:gap_i21k_vit} shows that a ViT architecture trained on ImageNet21K exhibits the reverse trend SGD$>$Adam, with a gap of $\Delta^*=+0.851$, or $\approx12\%$ in favor of SGD (see~\Cref{app:exp_i21k} for experimental details). 
While adaptivity in the head and embedding layer narrows the gap, other blocks may also play a role. To understand these effects, we next examine how progressive architectural changes shift the optimizer preference.

\input{tables/arch_interpolation}
\textbf{From SGD to Adam via architecture interpolation.}
We next test whether architectural design can shift optimizer preference on a fixed dataset.
Training on ImageNet21K, we progressively introduce ConvNeXt-style~\citep{liu2022convnet} design choices, inspired by Vision Transformers, to ResNet50~\citep{he2016deep}, demonstrating that a few simple architecture changes can interpolate the gap from SGD to Adam advantage. The following three architecture choices are sufficient: (1) BatchNorm $\rightarrow$ LayerNorm; (2) ReLU $\rightarrow$ GeLU; (3) Dense $\rightarrow$ Depthwise convolution. Further experimental details are reported in~\Cref{app:exp_arch_interpolation}.

% Results
\Cref{tab:arch_interpolation} shows the gap for one-epoch training on I21K. Starting from the ResNet50, SGD is $+13.00\%$ faster than Adam. Stacking each architecture change on top of another, the gap $\Delta$ shifts from SGD to Adam advantage as follows: (1) +~LayerNorm$=+7.29\%$; (2) +~GeLU$=-0.49\%$; (3) +~Depthwise convolution$=-11.86\%$. Finally, the original ConvNext has a $\Delta=-15.53\%$. This interpolation exercise highlights how the architectural design is coupled with the optimizer choice (the effects of each modification independently are reported  in~\Cref{app:exp_arch_interpolation_independent}). While architecture can shift the advantage between optimizers on a fixed dataset, it does not determine a universal winner. This motivates our next investigation beyond data and architecture.

\subsection{Batch size and the number of steps}
\label{subsec:exp_opt}

Previous sections found that the gap cannot be fully explained by single-factor explanations alone. Here, we connect these explanations from yet another angle: \emph{batch size}, showing how both data and architecture properties can shift the advantage between SGD and Adam.

\textbf{Batch size.} In the Transformer LM setup, recent works have shown that the Adam--SGD gap nearly vanishes at small batch sizes~\citep{marek2025small} and can even reverse to an SGD advantage~\citep{sreckovic2025your}. At first glance, this seems yet another explanation. Does it connect with previous data and architecture hypotheses? Does the same finding transfer to other domains? We answer these questions by broadening the prior batch size studies to more setups, including additional data domains (vision, graphs, and genomics) and classic non-Transformer architectures. 
\Cref{fig:main} summarizes the results across a wide range of experimental setups at different batch sizes $B$. The SGD to Adam shift presented in prior works generalizes across all our setups, though the rate at which the advantage shifts varies by configuration.
For example, ViT trained on I21K has a steeper slope compared to GPT trained on FineWeb or HG38, with Graph Transformer (GRIT) sitting in between. Similar trends hold for non-Transformer architectures such as GCNN, ResNet, and GAT trained on the same datasets. Moreover, models trained on I21K have the largest \emph{crossover batch size} across all the settings, regardless of the architecture family. This is also consistent with the strong empirical performance of SGD in vision tasks compared to language modeling~\citep{kasimbeg2025accelerating}.  
\emph{These results suggest that both data and architecture determine how rapidly, and at what batch size, the advantage shifts from SGD to Adam.}

A natural question arises from the previous finding, which sweeps the batch size under a fixed number of training samples. This implies that a smaller $B$ takes more steps than a larger one. Is the number of steps a possible confounder?~\citet{sreckovic2025your} show that under a simple Gaussian-noisy quadratic model, the early training phase of SGD is $B$-independent and driven mainly by the number of steps, whereas SignSGD gains a $\smash{\sqrt{B}}$ speedup factor up to the critical batch size~\citep{compagnoni2025adaptive}. This suggests that adaptive methods' advantage may be explained by quick early training progress. 
We revisit this below, showing the gap is pronounced in the modern underparametrized regime (one-epoch training), as opposed to the classical overparametrized setting (multi-epoch), where optimization can reach the minimizer~\citep{belkin2019reconciling}.

\begin{figure}[h]
\vspace{-2mm}
  \centering
  \begin{minipage}[c]{0.50\linewidth}
  \begin{subfigure}[c]{.52\linewidth}
    \centering
    \includegraphics[width=\linewidth]{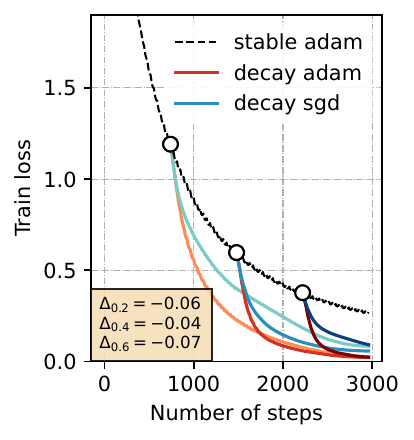}
    \caption{Overparametrized}
  \end{subfigure}%
  \begin{subfigure}[c]{.465\linewidth}
    \centering
    \includegraphics[width=\linewidth]{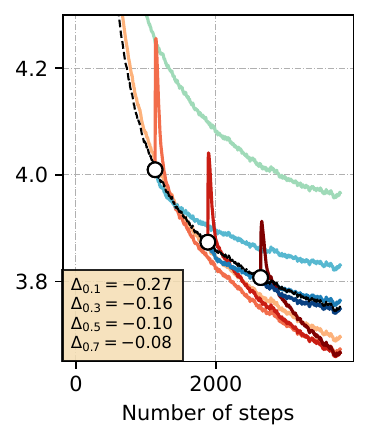}
    \caption{Underparametrized}
  \end{subfigure}
  \end{minipage}
  \hfill
  \begin{minipage}[c]{0.48\linewidth}
    \centering
    \caption{GPT 50M parameters trained on (a) 10M TinyStories tokens for 80 epochs and (b) 1B FineWeb tokens for 1 epoch at $B=1024$. 
    At different stages of training, we isolate the effect of early training dynamics by branching Adam/SGD from a shared constant learning rate trajectory (\texttt{stable\_adam}). 
    \textbf{(a)} For overparametrization, both branches converge to low loss as the problem is solvable with more epochs, while \textbf{(b)} for underparametrization, SGD lags behind.
    For every branch, the learning rate and momentum are tuned, see~\Cref{app:exp_early_training}.
    }
    \label{fig:early_training}
  \end{minipage}
\end{figure}
\vspace{-2mm}
\textbf{Early training phase.} 
In classical optimization, overparameterization refers to the interpolation regime: the model is expressive enough to fit the training set perfectly~\citep{vaswani2020adaptive,loizou2021stochastic}.
We adopt the term in an operational sense to distinguish between LMs trained for many epochs on a small dataset (overparametrized) from those trained for one epoch on a large dataset (underparametrized). Under this view, with step count and batch size fixed, ``earliness'' in training is inversely proportional to the number of steps performed. We design an experiment to isolate the \emph{early training phase} and measure the Adam--SGD gap in both regimes.
We use GPT in two scenarios: (a) an \emph{overparametrized} setup, training on 10M TinyStories tokens for multiple epochs until train loss $\approx 0$; and (b) an \emph{underparametrized} setup, training on 1B FineWeb tokens for one epoch. Following the WSD scheduler~\citep{hu2024minicpm}, we first train a shared trajectory \texttt{stable\_adam} with a constant learning rate, saving multiple checkpoints along training to isolate different stages of the early phase. From each checkpoint, we branch and continue training with either Adam or SGD with a linear decay schedule to 0. For a fair comparison, the momentum state is reset at each branch point, with $\eta$ and $\beta$ extensively tuned, see~\Cref{app:exp_early_training}.

% Results
\Cref{fig:early_training} shows the Adam--SGD gap in both scenarios. In the overparametrized setting, we observe that \texttt{decay\_sgd} is slower but is able to reach train loss $\approx$ 0, similar but slower than Adam. For underparametrization, \texttt{decay\_sgd} can only approximately match the \texttt{stable\_adam} curve, while \texttt{decay\_adam} consistently reaches a lower loss. These two scenarios present a striking contrast that bridges the classic overparametrized literature and the modern literature dominated by Transformer-based LMs. 
This experiment suggests that the early training hypothesis is better suited to describe an overparametrized setup. In Appendix~\Cref{app:exp_steps}, we present a preliminary study on the specific role of the number of steps.

\section{Theoretical model of the gap}
\label{sec:theory}

We study how the Adam--SGD gap depends on batch size through the nonconvex bounds of \citet{kovalev2025understanding}. These bounds have been shown to predict optimizer hyperparameter scaling with the token budget $T$ \citep{shulgin2026deriving,islamov2026role}.

\subsection{Convergence bounds}

We consider the stochastic optimization problem
\begin{equation}\label{eq:problem}
    \min_{x\in\R^d} \left[f(x)=\E_{\xi\sim\gD}[f(x,\xi)]\right],
\end{equation}
and compare NSGD with momentum and SignSGD with momentum in solving Equation~\ref{eq:problem}. These methods serve as theoretical proxies for studying clipped SGD and Adam, respectively.

Let $f$ be $L$-smooth with respect to a norm $\|\cdot\|$ and the mini-batch gradient noise variance be bounded by $\sigma^2/B$. In the regime $T\gg1$, with tuned learning rate and momentum parameters, \citet{kovalev2025understanding} obtain the convergence bound 
\[
U_{\|\cdot\|} \propto
\frac{(L\delta_0)^{1/4}\sqrt{\rho\sigma}}{T^{1/4}}
+
\frac{(\rho\sigma)^2}{\sqrt{L\delta_0TB}}
+
\frac{(L\delta_0)^{3/4}B}{\sqrt{\rho\sigma}T^{3/4}},
\]
where $U_{\|\cdot\|}$ denotes the convergence measure corresponding to the norm $\|\cdot\|$ and $\delta_0=f(x_0)-f^*$ is the initial suboptimality gap. Specializing this bound to NSGD corresponds to choosing the $\smash{\ell_2}$ norm, with $L=L_2$ and $\rho=1$. For SignSGD, we choose the $\ell_\infty$ norm, with $L=L_\infty$ and $\smash{\rho=\sqrt d}$, where $d$ is the problem dimension.

\subsection{Convergence gap regimes depend on the batch size}

\begin{figure}[h]
  \centering
  \begin{minipage}[c]{0.55\linewidth}
    \centering
    \footnotesize
    \label{tab:regimes}
    \begin{tabular}{ccl}
    \toprule
    \textbf{Regime} & \textbf{Condition} & \textbf{Better optimizer across $B$} \\
    \midrule
    I & $1 < r < d^{1/3}$ & SignSGD everywhere, or\\
     & & NSGD to SignSGD \emph{crossover} \\
    II & $d^{1/3} < r < d$ & NSGD everywhere \\
    \bottomrule
    \end{tabular}

    \vspace{1.5em}

    \begin{subfigure}[c]{.45\linewidth}
      \centering
      \includegraphics[width=\linewidth]{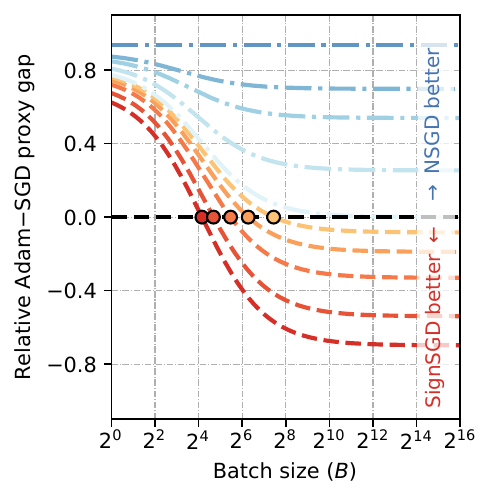}
    \end{subfigure}%
    \begin{subfigure}[c]{.55\linewidth}
      \centering
      \includegraphics[width=\linewidth]{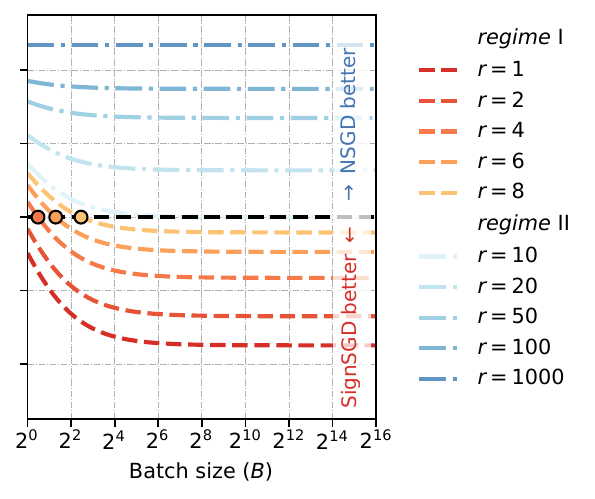}
    \end{subfigure}
  \end{minipage}
  \hfill
  \begin{minipage}[c]{0.425\linewidth}
    \vspace{-1em}
    \caption{Theoretical model for the Adam--SGD gap. \textbf{(top)} The two gap regimes as a function of $r$. \textbf{(bottom)} Relative gap $\left(U_{\infty}(B) - U_{2}(B)\right) / \left(U_{\infty}(B) + U_{2}(B)\right)$ where $U_{\infty}$ is the {\setlength{\fboxsep}{1pt}{\colorbox{red!20}{SignSGD}}} bound and $U_2$ the {\setlength{\fboxsep}{1pt}{\colorbox{Cerulean!20}{NSGD}}} bound. 
    \emph{Regime I} curves show a crossover from NSGD to SignSGD in the left panel and favor SignSGD for all B in the right panel; 
    \emph{Regime II} curves consistently favor NSGD for all $B$. Circles mark the crossover points.
    The gap depends on the $\Delta_U$'s constants, which we fix at $d=10^3$, $T=10^6$, $\delta_0=10$, $\sigma=0.1$.
    The two panels differ only in $L_2$ (left $L_2=1$; right $L_2=1000$). 
    }
    \label{fig:theory}
  \end{minipage}
\end{figure}

We define the gap between the SignSGD and NSGD bounds as $\Delta_U=U_\infty - U_2$, which expands to 
\begin{equation}\label{eq:gap}
    \Delta_U
    =
    \frac{(\sigma^2\delta_0)^{1/4}\left((L_\infty d)^{1/4}-L_2^{1/4}\right)}{T^{1/4}}
    +
    \frac{\sigma^2}{\sqrt{\delta_0TB}}\left(\frac{d}{\sqrt{L_\infty}}-\frac{1}{\sqrt{L_2}}\right)
    +
    \frac{\delta_0^{3/4}B}{\sqrt{\sigma}T^{3/4}}\left(\frac{L_\infty^{3/4}}{d^{1/4}}-L_2^{3/4}\right). \notag
\end{equation}
When $\Delta_U>0$ NSGD is favored, while $\Delta_U<0$ favors SignSGD. Let $r=L_\infty/L_2$ be the geometry smoothness ratio, then we always have the relation $1 \le r \le d$, since $L_2 \le L_\infty \le dL_2$. Depending on the value of $r$, the gap has two possible regimes summarized in~\Cref{fig:theory}, with the Appendix~\Cref{app:theory} reporting the complete derivations.

In regime I, the derivative of the gap $\Delta_U$ with respect to $B$ is always negative. Therefore, two outcomes are possible. If $\Delta_U(1)>0$, NSGD achieves the lower bound at small batch sizes, but increasing $B$ eventually produces a crossover beyond which SignSGD is favored. If $\Delta_U(1)<0$, SignSGD achieves the lower bound already at $B=1$ and remains favored for all $B$s.
In Regime II, the gap $\Delta_U>0$ for all $B$s, indicating that NSGD always achieves the lower convergence bound.
The geometry ratio $r=L_\infty/L_2$ determines which regime applies, while the remaining problem-dependent quantities $T,\sigma,\delta_0$ determine whether Regime I exhibits a crossover or favors SignSGD at different $B$s. This predicts that the relative performance of Adam and SGD depends jointly on the optimization geometry, dataset, and gradient noise characteristics.

Connecting to our numerical results in the previous sections, most of those experimental setups fall under the Regime I with a crossover point from SGD to Adam advantage as $B$ scales. We also observe two additional setups covering the remaining cases. For the GPT architecture trained on genomics data HG38, Adam$>$SGD for all the batch sizes starting from $B=1$ (see~\Cref{app:bs1}). This corresponds to the Regime I without the crossover. For the ResNet50 architecture on I21K, SGD$>$Adam even at large $B=65536$ (see~\Cref{app:exp_steps}), corresponding to the Regime II. Thus, our experimental results match the different regimes identified with the proposed theoretical model.

\subsection{Power-law spectrum}

We now use a simple stochastic quadratic model to make the geometry ratio $r=L_\infty/L_2$ more explicit. This lets us relate the regimes above to the Hessian spectrum. The model we consider is
\[
f(x,\xi)
=
\frac{1}{2}x^\top \operatorname{diag}(\lambda_1,\dots,\lambda_d)x
+
\langle \xi,x\rangle,
\]
where $|\lambda_1|\geq |\lambda_2|\geq \cdots \geq |\lambda_d|$,
$\E[\xi]=0$, and $\E[\|\xi\|^2]\leq \sigma^2/B$. For this function,
$L_2=|\lambda_1|$, while
$
L_\infty=\sum_{i=1}^d |\lambda_i|.
$
To understand how the smoothness constants scale with dimension, we assume a power-law spectrum
$
|\lambda_j| \propto \frac{|\lambda_1|}{j^\beta}
$
\citep{kunstner2025scaling}. For $d\gg 1$, this gives
\[
L_\infty \propto |\lambda_1|\times
\begin{cases}
    d^{1-\beta}, & \text{if } \beta < 1,\\
    \log d, & \text{if } \beta=1,\\
    1, & \text{if } \beta > 1.
\end{cases}
\]
Equivalently, we have $r\propto d^{1-\beta}$ for $\beta<1$, $r\propto \log d$ for $\beta=1$, and $r\propto 1$ for $\beta>1$. Hence, if $\beta<2/3$, then $r\gtrsim d^{1/3}$, which corresponds to a regime in which NSGD is favored. In this sense, a relatively homogeneous Hessian spectrum can make NSGD preferable. By contrast, when $\beta>2/3$, we have $r\lesssim d^{1/3}$, so the preferred optimizer may depend on the batch size. In this case, SignSGD may be favored for all batch sizes, or a batch-size-dependent crossover may occur. Because our analysis is stochastic and takes into account not only the Hessian heterogeneity but also the gradient noise, this behavior differs from the deterministic-regime conclusions of \citet{kunstner2025scaling}.

This example illustrates that Hessian heterogeneity plays an important role in the performance of an optimizer. At the same time, the best optimizer is not determined by the spectrum alone: it also depends on the batch size and the stochastic gradient noise, as discussed in the previous section.

% ----------------------------------------------------

\section{Discussion}
\label{sec:discussion}

We presented a broad empirical study on the Adam--SGD gap, pushing beyond the typical Transformer-based LM setup while carefully controlling for confounders. 
We revisited the existing single-factor gap explanations, 
% such as heavy-tailed vocabulary imbalance, softmax-attention, architecture component heterogeneity, and clarified differences between overparametrized and underparametrized regimes. 
% Our main result offers 
while offering a unified perspective centered on a \emph{crossover batch size} shaped by both data and architecture properties. This is supported by our theoretical model, which captures this batch-size dependence. 
Lastly, we observe that Adam is more stable under our joint hyperparameter sweeps~(\Cref{fig:fineweb_train_loss,fig:fineweb_gdn_train_loss,fig:i21k_train_loss,fig:zinc_train_loss,fig:hg38_char_train_loss,fig:fineweb_gcnn_char_train_loss}), leading to a consistent practical advantage.

Looking forward, our results show that every dataset-architecture pair has a different \emph{crossover} slope and point (\Cref{fig:main}).
This perspective suggests several follow-up questions: 
What data properties beyond vocabulary imbalance drive the Adam--SGD gap?
How do specific architectural choices modulate the crossover point?
And can this crossover inform the design of novel optimizers?

\textbf{Limitations.}
Although the Adam--SGD gap is a central question in modern large-scale optimization, recent practice has also reported promising results for newer optimizers, including Muon~\citep{jordan2024muon}, Scion~\citep{pethick2025training}, and SOAP~\citep{vyas2025soap}. Evaluating these methods in our setting is an important future direction; the literature is still evolving, with large-scale reports~\citep{shah2025practical, team2025kimi, deepseekai2025deepseekv4} using distinct variants of Muon while the algorithm itself continues to be refined theoretically~\citep{amsel2025polar}. Notably,~\citet{marek2025small} already show that Muon is effective in the large batch regime, suggesting that trends and conclusions in this work may generalize.
Lastly, the numerical experiments depend on the coverage of the hyperparameter grid, and the choice of the datasets and architectures. 

% \newpage
\section*{Acknowledgments}
Chenxiang Zhang acknowledges the financial support of the University of Luxembourg.
Rustem Islamov, Enea Monzio Compagnoni, and Aurelien Lucchi acknowledge the financial support of the Swiss National Science Foundation, SNSF grant No 207392.
Antonio Orvieto acknowledges the financial support of the Hector Foundation and the AI2050 Early Career Fellowship from Schmidt Sciences. We acknowledge computing resources provided by MPI-IS, MeluXina\footnote{\url{https://www.luxprovide.lu/meluxina/}}, and sciCORE\footnote{\url{http://scicore.unibas.ch/}}.

\section*{Author contributions}
\label{sec:contributions}
Chenxiang Zhang led the project, contributing to the experiments, infrastructure, figures, framing, direction, and writing. 
Rustem Islamov contributed to the experiments, framing, direction, writing, and theoretical result.
Enea Monzio Compagnoni and Jun Pang contributed to the framing and writing.
Aurelien Lucchi contributed to the framing, direction, and writing.
Antonio Orvieto is the lead advisor, contributing substantially to the framing, direction, writing, and theoretical result. 

%%%%%%%%%%%%%%%%%%%%%%%%%%%%%%%%%%%%%%%%%%%%%%%%%%%%%%%%%%%%

% \newpage
\bibliographystyle{plainnat}
\bibliography{reference}

\newpage
\section*{Appendix contents}
\startcontents[appendices]
\printcontents[appendices]{}{1}{\setcounter{tocdepth}{3}}
\appendix
\newpage

\section{Experimental details}
\label{app:exp_setup}

We present details for all the experiments in the main paper. Each run is performed on a single A100--40G NVIDIA GPU machine. 

\input{tables/arch}

\subsection{Datasets}

\paragraph{TinyStories.} Synthetic dataset of short stories that contain only simple words, generated by GPT-3.5 and GPT-4. We use the first 10M tokens for overparametrized settings, preprocessed with \texttt{gpt2} tokenizer.

\paragraph{FineWeb.} Large dataset derived from 96 Common Crawl snapshots. We use the first 1B tokens from the \texttt{fineweb-edu-sample-10BT}, preprocessed with \texttt{gpt2} tokenizer. 

\paragraph{I21K (ImageNet21K).} Large-scale image classification dataset containing approximately 14 million images across 21,841 synset classes from the ImageNet hierarchy. We use the \texttt{winter21} version: \url{https://huggingface.co/datasets/timm/imagenet-w21-wds} .

\paragraph{I1K heavy-tailed (ImageNet1K).} Subset of $n=10217$ samples from the ImageNet1K with a heavy-tailed class imbalance, constructed by sorting classes by frequency and sampling $\left\lceil \frac{1300}{k} \right\rceil$ images from the k-th class~\citep{kunstner2024heavy}.

\paragraph{HG38.} Human genome reference assembly GRCh38 (hg38), consisting of 2.7B nucleotides. Commonly used for training genomic sequence models. We use the preprocessed version: \url{https://storage.googleapis.com/basenji_barnyard2/hg38.ml.fa.gz} .

\paragraph{ZINC250K.} Subset of 250,000 drug-like molecules drawn from the ZINC database of commercially available compounds. Molecules are represented as SMILES strings and used as a benchmark for molecular generation and optimization tasks.

\subsection{Architectures}
\label{app:arch}

\paragraph{GPT: Transformer for language tasks.}\label{par:language_transformer_description}
For pre-training Transformers under the causal language modeling objective, we build on the nanoGPT implementation \citep{karpathy2022nanogpt}, extending it with several recent model advances:
\begin{itemize}[leftmargin=*]
    \item Causal scaled dot-product attention
    \item RoPE \citep{su2024roformer}
    \item PreNorm \citep{xiong2020layer}
    \item RMSNorm \citep{zhang2019root}
    \item QKNorm \citep{touvron2023llama}
    \item EmbedNorm \citep{loshchilov2025ngpt}
    \item ReLU${}^2$ MLP with ratio 4 \citep{so2021searching}
    \item No biases
    \item Linear head
\end{itemize}
The configuration:
\begin{itemize}[leftmargin=*]
    \item Layers: 6
    \item Heads: 6
    \item Embedding size: 384
    \item Sequence length: 256
    \item Vocabulary size: [data dependent]
\end{itemize}

\paragraph{ViT: Transformer for vision tasks.}\label{par:vision_transformer_description}
We follow and extend the original model from~\citep{dosovitskiy2020image} with recent model advances:
\begin{itemize}[leftmargin=*]
    \item Scaled dot-product attention
    \item Conv2D stem~\citep{xiao2021early}  \item Mean pooling
    \item RoPE2D axial / sincos2D \citep{heo2024rotary}
    \item PreNorm \citep{xiong2020layer}
    \item RMSNorm \citep{zhang2019root}
    \item QKNorm \citep{touvron2023llama}
    \item EmbedNorm \citep{loshchilov2025ngpt}
    \item ReLU${}^2$ MLP with ratio 4 \citep{so2021searching}
    \item No biases
    \item Linear head
\end{itemize}
The configuration:
\begin{itemize}[leftmargin=*]
    \item Layers: 6
    \item Heads: 6
    \item Embedding size: 384
    \item Patch size: 16
    \item Vocabulary size: [data dependent]
\end{itemize}

\paragraph{GPT: Transformer for genomic tasks.}
The model is the same as the GPT for language tasks, but with a smaller vocabulary size depending on the tokenizer \{\texttt{char}, \texttt{kmer}\}. The model configuration is the same as the GPT for language, except for a sequence length of 1024 and a smaller vocabulary.

\paragraph{GRIT: Transformer for graph tasks.}
We use the implementation from the original publicly released code: \url{https://github.com/LiamMa/GRIT/tree/main}.

This configuration has 0.27M parameters:
\begin{itemize}[leftmargin=*]
    \item Layers: 6
    \item Heads: 6
    \item Embedding size: 64
    \item $k$-steps: 21~\citep{ma2023grit}
\end{itemize}

\paragraph{ResNet50: Conv model for vision tasks.} We use timm's implementation: \texttt{resnet50.a1\_in1k}

\paragraph{ConvNext: Conv model for vision tasks.} We use timm's implementation: \texttt{convnext\_tiny}

\paragraph{GCNN: Conv model for language tasks.}
This revamped version of the gated convolution architecture is identical to the GPT for language tasks except for two changes:
\begin{itemize}[leftmargin=*]
    \item Softmax-attention is replaced by the GLU layer: $h(X) = (X * W + b) \otimes \sigma(X * V + c)$
    \item Sinusoidal positional embedding
\end{itemize}
The configuration is identical to the GPT model.

\paragraph{GAT: Message-passing model for graph tasks.}
Graph attention network is the model used for graph-level regression \citep{velivckovic2018graph}. The model architecture is: 
\begin{itemize}[leftmargin=*]
    \item Embedding layer
    \item GATConv $\rightarrow$ Batch norm $\rightarrow$ ReLU blocks
    \item Global mean pooling
    \item Linear head
\end{itemize}
This configuration has 0.27M parameters:
\begin{itemize}[leftmargin=*]
    \item Layers: 4
    \item Heads: 8
    \item Embedding size: 256
\end{itemize}

\paragraph{GDN: linear Transformer for language tasks.} We implement the GatedDeltaNet~\citep{yang2025gated} identical to the GPT architecture used for language tasks except for two changes:
\begin{itemize}[leftmargin=*]
    \item Softmax-attention is replaced by the GatedDeltaNet block\footnote{\url{https://github.com/fla-org/flash-linear-attention}}
    \item Positional embedding is removed
\end{itemize}
For genomic task, same as GPT, we use a sequence length of 1024 and smaller vocabulary with the \texttt{char} tokenizer.

% -------------------------------------------------------

\newpage
\subsection{Experiments on FineWeb}\label{app:exp_fineweb}
For the experiments in \Cref{fig:main,fig:gap_gcnn}, we use the model GPT 50M parameters (see~\Cref{app:arch}) trained on 1B tokens (Chinchilla-optimal ratio of 20 tokens/parameter).
We use \texttt{bfloat16} precision, 10\% linear warmup with a cosine scheduler to 0, and global gradient norm clipping to 1 for both Adam and SGD. 

We tune the learning rate $\eta$, momentum $\beta$, and batch size $B$ as
\begin{itemize}[leftmargin=*]
    \item SGD (96 runs)
        \begin{align*}
        (\eta, \beta, B) &\in \{10^{-5}, 10^{-4}, 10^{-3}, 10^{-2}, 10^{-1}, 10^0\}\\
                      &\times \{0.9, 0.95, 0.99, 0.999\}\\
                      &\times \{16, 64, 256, 1024\}.
        \end{align*}
    \item Adam (96 runs) ($\beta_1=\beta_2=\beta$) 
        \begin{align*}
        (\eta, \beta, B) &\in \{10^{-4}, 10^{-3}, 10^{-2}, 10^{-1}, 10^0, 10^{1}\}\\
                      &\times \{0.9, 0.95, 0.99, 0.999\}\\
                      &\times \{16, 64, 256, 1024\}.
        \end{align*}
\end{itemize}
For $B=16$, we perform a more finegrained sweep by reducing the edge of the grid and adding more points inbetween, adding 8 more configurations.

This setup trains a total of 200 model configurations.

\begin{figure}[!ht]
  \centering
  \includegraphics[width=\linewidth]{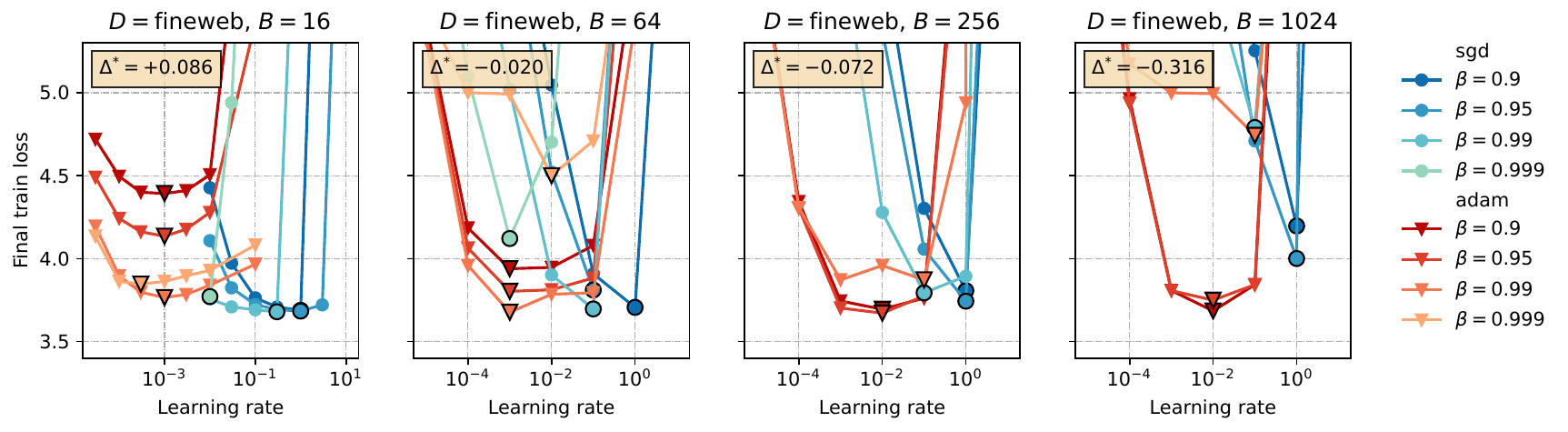}
  \caption{GPT 50M parameters trained on FineWeb 1B tokens. 
  }
  \label{fig:fineweb_train_loss}
\end{figure}

\begin{figure}[!ht]
  \centering
  \includegraphics[width=0.8\linewidth]{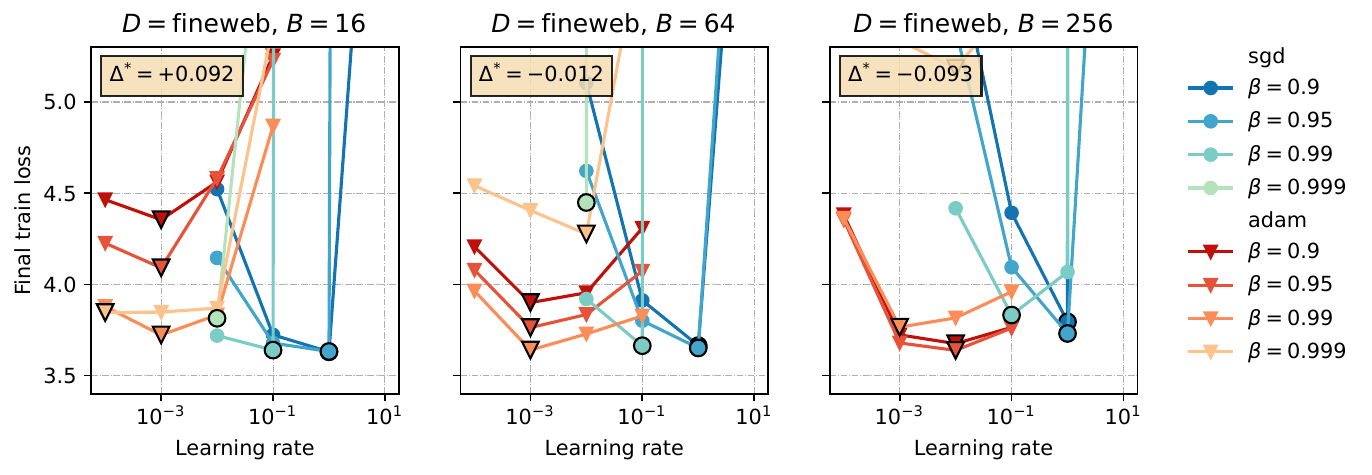}
  \caption{GDN 50M parameters trained on FineWeb 1B tokens. 
  }
  \label{fig:fineweb_gdn_train_loss}
\end{figure}

\newpage
\subsection{Experiments on ImageNet21K}\label{app:exp_i21k}
For the experiments in \Cref{fig:main}, we use ViT and ResNet50 models (see~\Cref{app:arch}) trained for 1 epoch on ImageNet21K. We use \texttt{bfloat16} precision, 10\% linear warmup with cosine scheduler to 0, and global norm clipping to 1 for both Adam and SGD.

\textbf{ViT.} We tune the learning rate $\eta$, momentum $\beta$, and batch size $B$ as
\begin{itemize}[leftmargin=*]
    \item SGD (48 runs)
        \begin{align*}
        (\eta, \beta, B) &\in \{10^{-2}, 10^{-1}, 10^{0}, 10^{1}\}\\
                      &\times \{0.9, 0.95, 0.99, 0.999\}\\
                      &\times \{256, 1024, 4096\}.
        \end{align*}
    \item Adam (48 runs) ($\beta_1=\beta_2=\beta$) 
        \begin{align*}
        (\eta, \beta, B) &\in \{10^{-4}, 10^{-3}, 10^{-2}, 10^{-1}\}\\
                      &\times \{0.9, 0.95, 0.99, 0.999\}\\
                      &\times \{256, 1024, 4096\}.
        \end{align*}
\end{itemize}

\textbf{ResNet50.} We tune the learning rate $\eta$, momentum $\beta$, and batch size $B$ as
\begin{itemize}[leftmargin=*]
    \item SGD (48 runs)
        \begin{align*}
        (\eta, \beta, B) &\in \{10^{-2}, 10^{-1}, 10^{0}, 10^{1}\}\\
                      &\times \{0.9, 0.95, 0.99, 0.999\}\\
                      &\times \{1024, 4096, 16384\}.
        \end{align*}
    \item Adam (48 runs) ($\beta_1=\beta_2=\beta$) 
        \begin{align*}
        (\eta, \beta, B) &\in \{10^{-4}, 10^{-3}, 10^{-2}, 10^{-1}\}\\
                      &\times \{0.9, 0.95, 0.99, 0.999\}\\
                      &\times \{1024, 4096, 16384\}.
        \end{align*}
\end{itemize}
This setup trains a total of 192 model configurations.

\begin{figure}[!ht]
  \centering
  \includegraphics[width=0.8\linewidth]{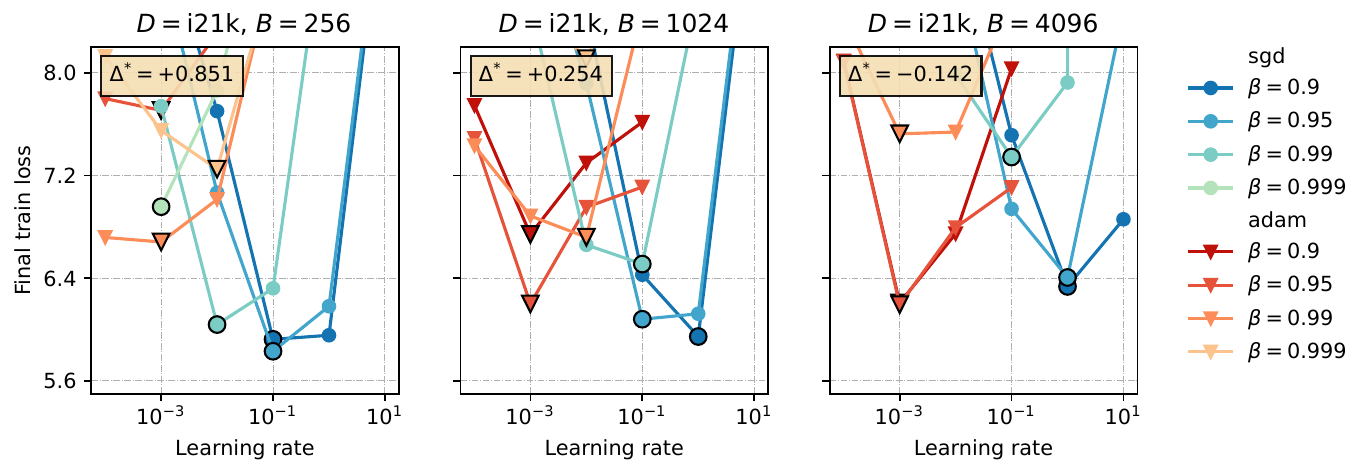}
  \includegraphics[width=\linewidth]{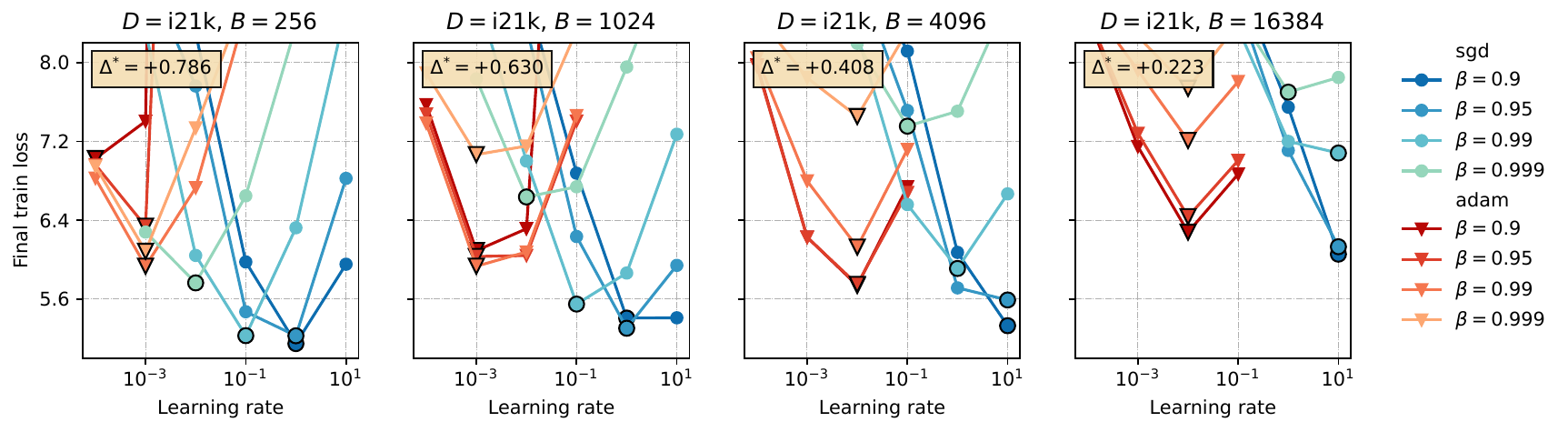}
  \caption{ViT (top) and ResNet50 (bottom) trained on I21K (13M training samples). 
  }
  \label{fig:i21k_train_loss}
\end{figure}

\newpage
\subsection{Experiments on ImageNet1K with heavy-tail}\label{app:exp_ht_i1k}
For the experiments in~\Cref{fig:gap_i1k_imb}, we reproduce the~\citet{kunstner2024heavy}'s results using their SimpleViT architecture from \texttt{\url{https://github.com/lucidrains/vit-pytorch}} \texttt{(version=1.6.5)} compared to our modernized ViT architecture~(see \Cref{app:arch}). Both models have 12 layers and are trained for 300 epochs with a constant learning rate, flip and random augmentations, $B=1024$, sinusoidal position embedding, and fixed $\beta_2=0.999$. 
We use \texttt{bfloat16} precision. We apply global gradient norm clipping to 1 to our ViT model in~\Cref{fig:gap_i1k_imb_vit}.

We tune the learning rate $\eta$, momentum $\beta$, and batch size $B$ as
\begin{itemize}[leftmargin=*]
    \item SGD (28 runs)
        \begin{align*}
        (\eta, \beta, B) &\in \{10^{-3}, 3\cdot10^{-3}, 10^{-2}, 3\cdot10^{-2}, 10^{-1}, 3\cdot10^{-1}, 10^0\}\\
                      &\times \{0.9, 0.95, 0.99, 0.999\}\\
                      &\times \{1024\}.
        \end{align*}
    \item Adam (28 runs) ($\beta_2=0.999$): 
        \begin{align*}
        (\eta, \beta_1, B) &\in \{10^{-5}, 3\cdot10^{-5}, 10^{-4}, 3\cdot10^{-4}, 10^{-3}, 3\cdot10^{-3}, 10^{-2}\}\\
                      &\times \{0.9, 0.95, 0.99, 0.999\}\\
                      &\times \{1024\}.
        \end{align*}
\end{itemize}
This setup trains a total of 112 model configurations, as there are two configurations.

\newpage
\subsection{Experiments on HG38}\label{app:exp_hg38}
For the experiments in~\Cref{fig:main,fig:gap_hg38}, we use the GPT model~(see \Cref{app:arch}) trained on the HG38 dataset preprocessed using three different tokenizers: \texttt{char} (10M parameters model on 2.7B tokens), \texttt{4mer} (10M parameters model on 0.68B tokens), and \texttt{6mer} (13M parameters model 0.45B tokens). 
We use \texttt{bfloat16} precision. We apply global gradient norm clipping to 1 for both Adam and SGD.

\textbf{Tokenizer \texttt{char}.} We tune the learning rate $\eta$, momentum $\beta$, and batch size $B$ as
\begin{itemize}[leftmargin=*]
    \item SGD (96 runs)
        \begin{align*}
        (\eta, \beta, B) &\in \{10^{-2}, 3\cdot10^{-2}, 10^{-1}, 3\cdot10^{-1}, 10^0, 3\cdot10^0\}\\
                      &\times \{0.9, 0.95, 0.99, 0.999\}\\
                      &\times \{16, 64, 256, 1024\}.
        \end{align*}
    \item Adam (96 runs) ($\beta_1=\beta_2=\beta$) 
        \begin{align*}
        (\eta, \beta, B) &\in \{10^{-5}, 3\cdot10^{-5}, 10^{-4}, 3\cdot10^{-4}, 10^{-3}, 3\cdot10^{-3}\}\\
                      &\times \{0.9, 0.95, 0.99, 0.999\}\\
                      &\times \{16, 64, 256, 1024\}.
        \end{align*}
\end{itemize}

\textbf{Tokenizer \texttt{4mer} and \texttt{6mer}.} We tune the learning rate $\eta$, momentum $\beta$, with batch size $B=256$ as
\begin{itemize}[leftmargin=*]
    \item SGD (24 runs)
        \begin{align*}
        (\eta, \beta) &\in \{10^{-2}, 3\cdot10^{-2}, 10^{-1}, 3\cdot10^{-1}, 10^0, 3\cdot10^0\}\\
                      &\times \{0.9, 0.95, 0.99, 0.999\}.\\
        \end{align*}
    \item Adam (24 runs) ($\beta_1=\beta_2=\beta$) 
        \begin{align*}
        (\eta, \beta) &\in \{10^{-5}, 3\cdot10^{-5}, 10^{-4}, 3\cdot10^{-4}, 10^{-3}, 3\cdot10^{-3}\}\\
                      &\times \{0.9, 0.95, 0.99, 0.999\}.\\
        \end{align*}
\end{itemize}

\vspace{-1.5em}
\textbf{Tokenizer \texttt{char} with GDN.} Same setup as GPT on genomics but using the model GDN. We tune the hyperparams grid as GPT on genomics above. 

This setup trains a total of 432 model configurations.

\begin{figure}[!ht]
  \centering
  \includegraphics[width=\linewidth]{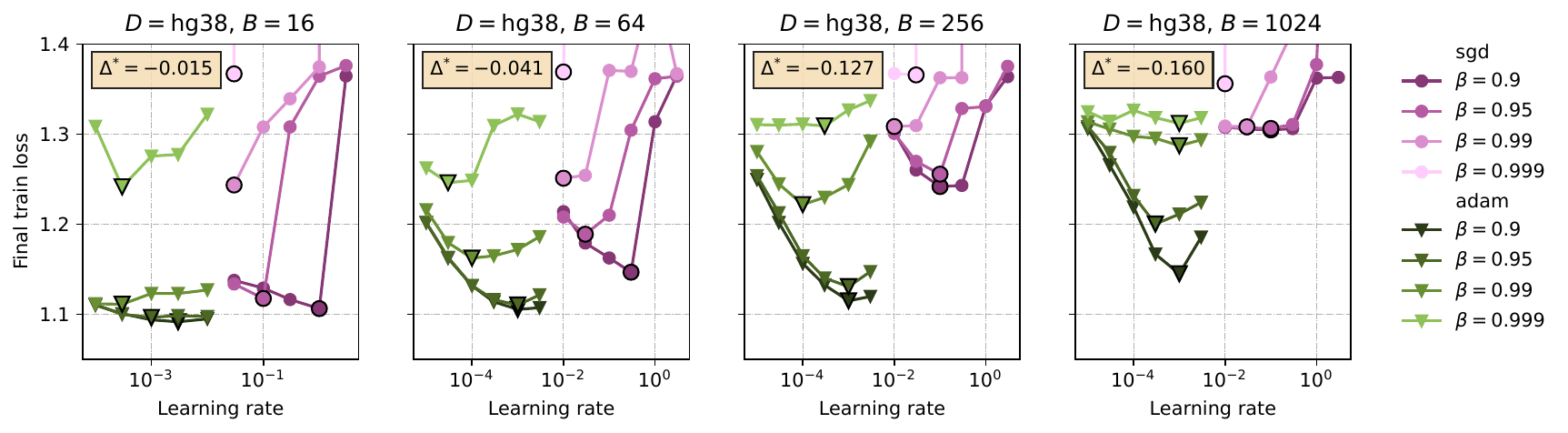}
  \includegraphics[width=\linewidth]{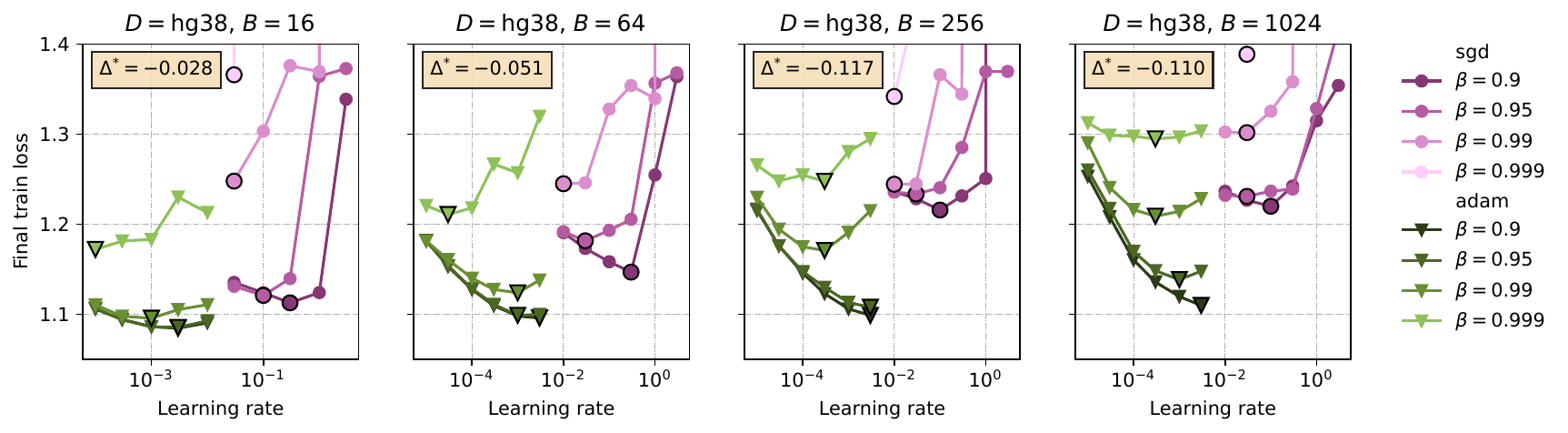}
  \caption{GPT (top) and GDN (bottom) 10M parameters trained on HG38 2.7B tokens.}
  \label{fig:hg38_char_train_loss}
\end{figure}

\newpage
\subsection{Experiments on ZINC250K}\label{app:exp_zinc}
For the experiments in~\Cref{fig:main,fig:gap_zinc250k_grit}, we use GRIT 0.27M parameters and GAT 0.27M model parameters (see~\Cref{app:arch}) trained for 100 epochs on ZINC250K. 
We use \texttt{bfloat16} precision, 10\% linear warmup with cosine scheduler to 0, and global norm clipping to 1 for both Adam and SGD.

\textbf{GRIT.} We tune the learning rate $\eta$, momentum $\beta$, and batch size $B$ as
\begin{itemize}[leftmargin=*]
    \item SGD (64 runs)
        \begin{align*}
        (\eta, \beta, B) &\in \{10^{-3}, 10^{-2}, 10^{-1}, 10^{0}\}\\
                      &\times \{0.9, 0.95, 0.99, 0.999\}\\
                      &\times \{16, 64, 256, 1024\}.
        \end{align*}
    \item Adam (64 runs) ($\beta_1=\beta_2=\beta$) 
        \begin{align*}
        (\eta, \beta, B) &\in \{10^{-4}, 10^{-3}, 10^{-2}, 10^{-1}\}\\
                      &\times \{0.9, 0.95, 0.99, 0.999\}\\
                      &\times \{16, 64, 256, 1024\}.
        \end{align*}
\end{itemize}
\textbf{GAT.} We tune the learning rate $\eta$, momentum $\beta$, and batch size $B$ as
\begin{itemize}[leftmargin=*]
    \item SGD (48 runs)
        \begin{align*}
        (\eta, \beta, B) &\in \{10^{-2}, 10^{-1}, 10^{0}, 10^{1}\}\\
                      &\times \{0.9, 0.95, 0.99, 0.999\}\\
                      &\times \{256, 1024, 4096\}.
        \end{align*}
    \item Adam (48 runs) ($\beta_1=\beta_2=\beta$)
        \begin{align*}
        (\eta, \beta, B) &\in \{10^{-4}, 10^{-3}, 10^{-2}, 10^{-1}\}\\
                      &\times \{0.9, 0.95, 0.99, 0.999\}\\
                      &\times \{256, 1024, 4096\}.
        \end{align*}
\end{itemize}
This setup trains a total of 224 model configurations.

\begin{figure}[!ht]
  \centering
  \includegraphics[width=\linewidth]{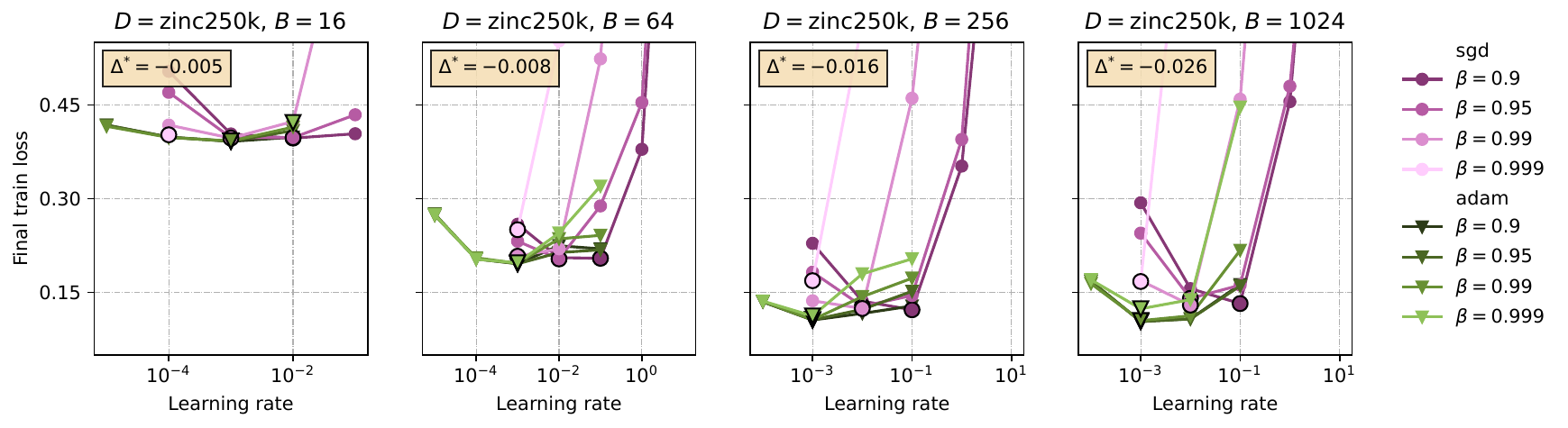}
  \includegraphics[width=0.8\linewidth]{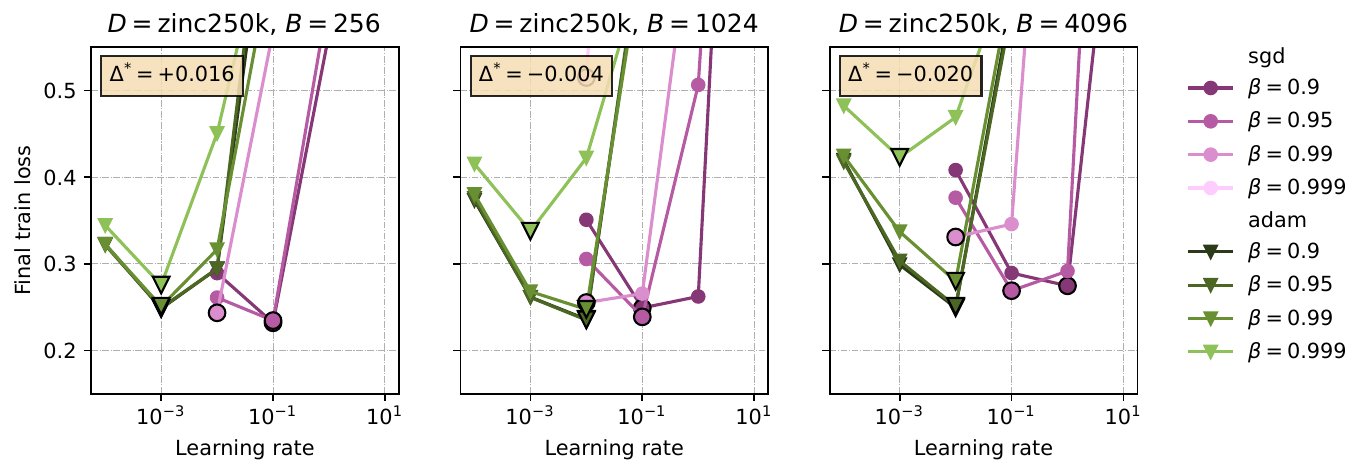}
  \caption{GRIT (top) and GAT (bottom) trained on ZINC250K. 
  }
  \label{fig:zinc_train_loss}
\end{figure}

\newpage
\subsection{Experiments on softmax-attention}\label{app:exp_gcnn}
For the experiments in \Cref{fig:main,fig:gap_gcnn}, we use the model GCNN 53M parameters (see~\Cref{app:arch}) trained on 1B tokens (Chinchilla-optimal 20 tokens/parameter).
We use \texttt{bfloat16} precision, 10\% linear warmup with cosine scheduler to 0, and global gradient norm clipping to 1 for both Adam and SGD. 

We tune the learning rate $\eta$, momentum $\beta$, and batch size $B$ as
\begin{itemize}[leftmargin=*]
    \item SGD (48 runs)
        \begin{align*}
        (\eta, \beta, B) &\in \{10^{-2}, 10^{-1}, 10^0, 10^1\}\\
                      &\times \{0.9, 0.95, 0.99, 0.999\}\\
                      &\times \{64, 256, 1024\}.
        \end{align*}
    \item Adam (48 runs) ($\beta_2=0.999$): 
        \begin{align*}
        (\eta, \beta_1, B) &\in \{10^{-4}, 10^{-3}, 10^{-2}, 10^{-1}\}\\
                      &\times \{0.9, 0.95, 0.99, 0.999\}\\
                      &\times \{64, 256, 1024\}.
        \end{align*}
\end{itemize}
For $B=1024$, an additional sweep is run for both SGD ($\eta=10^{-3}$) and Adam ($\eta=10^{0}$) to cover the edges induced by batch size scaling, resulting in 8 more runs.

This setup trains a total of 104 model configurations.

\begin{figure}[!ht]
  \centering
  \includegraphics[width=0.8\linewidth]{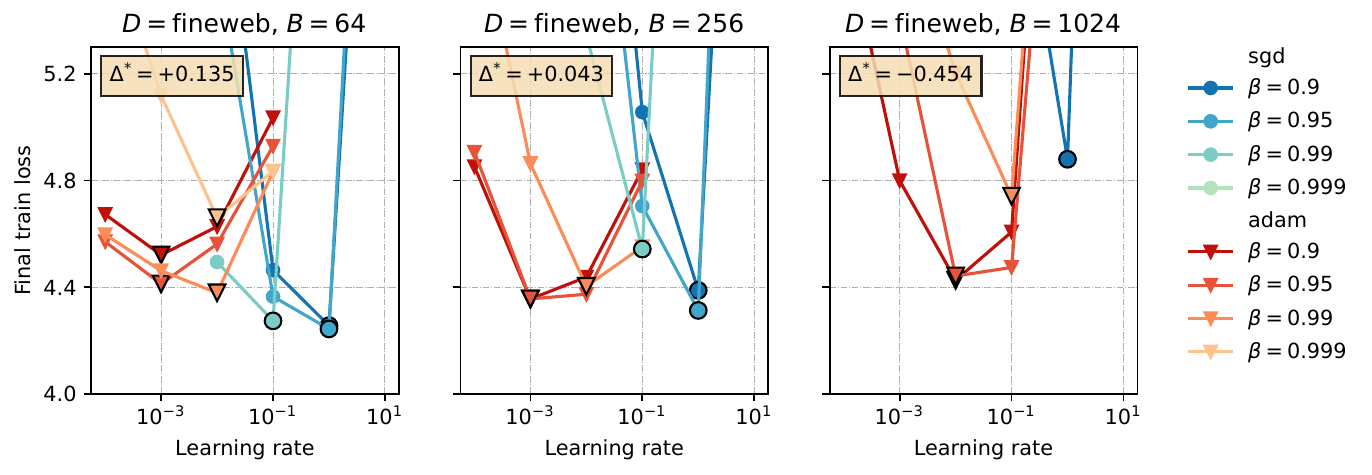}
  \caption{GCNN 53M parameters trained on FineWeb 1B. 
  }
  \label{fig:fineweb_gcnn_char_train_loss}
\end{figure}

\newpage
\subsection{Experiments on hybrid optimizers}\label{app:exp_hybrid}

For the experiments in \Cref{tab:hybrid_optimizer}, we use the model GPT 50M parameters (see~\Cref{app:arch}) trained on 1B tokens (Chinchilla-optimal 20 tokens/parameter).
In this set of experiments, particular layers are trained with Adam, while the rest of the network is trained with SGD. For both the Adam and SGD layers, we use \texttt{bfloat16} precision, 10\% linear warmup with cosine scheduler to 0, global gradient norm clipping to 1, and a batch size of 1024.

For each hybrid configuration of the optimizer, we tune the learning rates $\eta_{\text{SGD}}$ and $\eta_{\text{Adam}}$, momentum $\beta_{\text{SGD}}$ and $\beta_{\text{Adam}}$, and batch size $B$ as
\begin{itemize}[leftmargin=*]
    \item Hybrid optimizer (225 runs) (for Adam layers $\beta_1=\beta_2=\beta_{\text{Adam}}$)
        \begin{align*}
        (\eta_{\text{SGD}}, \eta_{\text{Adam}}, \beta_{\text{SGD}}, \beta_{\text{Adam}}, B) &\in \{10^{-3}, 10^{-2}, 10^{-1}, 10^{0}, 10^{1}\}\\
        &\times \{10^{-4}, 10^{-3}, 10^{-2}, 10^{-1}, 10^{0}\}\\
                      &\times \{0.9, 0.95, 0.99\}\\
                      &\times \{0.9, 0.95, 0.99\}\\
                      &\times \{1024\}.
        \end{align*}
\end{itemize}

This setup trains a total of 1125 model configurations, as there are 5 hybrid configurations in~\Cref{tab:hybrid_optimizer}.

\newpage
\subsection{Experiments on architecture interpolation}\label{app:exp_arch_interpolation}
For the experiments in \Cref{tab:arch_interpolation}, we use different variants of ResNet50 (see~\Cref{app:arch}) trained on I21K for one epoch.
We use \texttt{bfloat16} precision, 10\% linear warmup with cosine scheduler to 0, and global gradient norm clipping to 1 for both Adam and SGD. 

We keep the same ResNet50 structure through the interpolation, that is the same channel dimensions (256, 512, 1024, 2048) and depths (3, 4, 6, 3). Only specific layers are modified. The three modifications are:
\begin{itemize}[leftmargin=*]
    \item ReLU $\to$ GeLU. Substitute all ReLU with GeLU.
    \item BatchNorm $\to$ LayerNorm. Substitute all BatchNorm with LayerNorm.
    \item Dense mixing $\to$ Depthwise mixing. Each output channel only looks at its corresponding input channel. There is no cross-channel mixing as there is in Dense mixing~\citep{liu2022convnet}.
\end{itemize}

We tune the learning rate $\eta$, momentum $\beta$, and batch size $B$ as
\begin{itemize}[leftmargin=*]
    \item SGD (12 runs)
        \begin{align*}
        (\eta, \beta, B) &\in \{10^{-2}, 10^{-1}, 10^0\}\\
                      &\times \{0.9, 0.95, 0.99, 0.999\}\\
                      &\times \{1024\}.
        \end{align*}
    \item Adam (12 runs) ($\beta_1=\beta_2=\beta$): 
        \begin{align*}
        (\eta, \beta, B) &\in \{10^{-4}, 10^{-3}, 10^{-2}\}\\
                      &\times \{0.9, 0.95, 0.99, 0.999\}\\
                      &\times \{1024\}.
        \end{align*}
\end{itemize}

This setup trains a total of 72 model configurations, as there are 3 configurations in~\Cref{tab:arch_interpolation}.

\newpage
\subsection{Experiments on early training phase}\label{app:exp_early_training}
For the experiments in \Cref{fig:early_training}, we use GPT 50M parameters (see~\Cref{app:arch}) trained on 10M TinyStories tokens for 80 epochs and 1B FineWeb tokens for one epoch.
We use \texttt{bfloat16} precision, 10\% linear warmup with a constant learning rate to train the shared trajectory \texttt{stable\_adam}. For each branch, we reset the momentum state, tune the learning rate and momentum, apply the cosine scheduler to 0, and global gradient norm clipping to 1 for both Adam and SGD. 

There are 3 branching points for TinyStories at fractions \{0.2, 0.4, 0.6\} of the total training steps, while for FineWeb there are 4 checkpoints at \{0.1, 0.3, 0.5, 0.7\}. When branching in FineWeb, the batches that were already seen in the \texttt{stable} phase are correctly skipped. The total training budget is also reduced by skipping the \texttt{stable} steps.

We tune the learning rate $\eta$, momentum $\beta$, and batch size $B$ at each branch point as
\begin{itemize}[leftmargin=*]
    \item SGD (20 runs)
        \begin{align*}
        (\eta, \beta, B) &\in \{3\cdot10^{-2}, 10^{-1}, 3\cdot10^{-1}, 10^{0}, 3\cdot10^{0}\}\\
                      &\times \{0.9, 0.95, 0.99, 0.999\}\\
                      &\times \{1024\}.
        \end{align*}
    \item Adam (20 runs) ($\beta_1=\beta_2=\beta$): 
        \begin{align*}
        (\eta, \beta, B) &\in \{10^{-3}, 3\cdot10^{-3}, 10^{-2}, 3\cdot10^{-2}, 10^{-1}\}\\
                      &\times \{0.9, 0.95, 0.99, 0.999\}\\
                      &\times \{1024\}.
        \end{align*}
\end{itemize}

This setup trains a total of 280 model configurations, as there are 7 branches.

\newpage
\section{Additional results}

\subsection{Effect of weight decay}\label{app:exp_wd}

We ablate the effect of weight decay in both vision and language. The results are consistent with the trend observed in~\Cref{fig:main} without weight decay. That is, we still observe an optimizer shift from SGD to Adam as the batch size scales. However, the final train loss is higher than the corresponding run without weight decay. 

\paragraph{ViT on I21K.} As in~\Cref{app:exp_i21k}, we use a ViT model (see~\Cref{app:arch}) trained for 1 epoch on ImageNet21K. We use \texttt{bfloat16} precision, 10\% linear warmup with cosine scheduler to 0, and global norm clipping to 1 for both Adam and SGD. We use \emph{independent} weight decay~\citep{wortsman2024small}, which is decoupled and independent of learning rate $\eta$.

We tune the learning rate $\eta$, momentum $\beta$, and batch size $B$, also adding the weight decay $\lambda$ as
\begin{itemize}[leftmargin=*]
    \item SGD (48 runs)
        \begin{align*}
        (\eta, \beta, B, \lambda) &\in \{10^{-2}, 10^{-1}, 10^{0}, 10^{1}\}\\
                      &\times \{0.9, 0.95, 0.99\}\\
                      &\times \{1024, 4096\}\\
                      &\times \{10^{-3}, 10^{-4}\}.
        \end{align*}
    \item Adam (48 runs) ($\beta_1=\beta_2=\beta$) 
        \begin{align*}
        (\eta, \beta, B, \lambda) &\in \{10^{-4}, 10^{-3}, 10^{-2}, 10^{-1}\}\\
                      &\times \{0.9, 0.95, 0.99\}\\
                      &\times \{1024, 4096\}\\
                      &\times \{10^{-3}, 10^{-4}\}.
        \end{align*}
\end{itemize}
This setup trains a total of 120 model configurations. For $\lambda=10^{-3}$, we have 24 more runs.

\begin{figure}[!ht]
  \centering
  \includegraphics[width=0.8\linewidth]{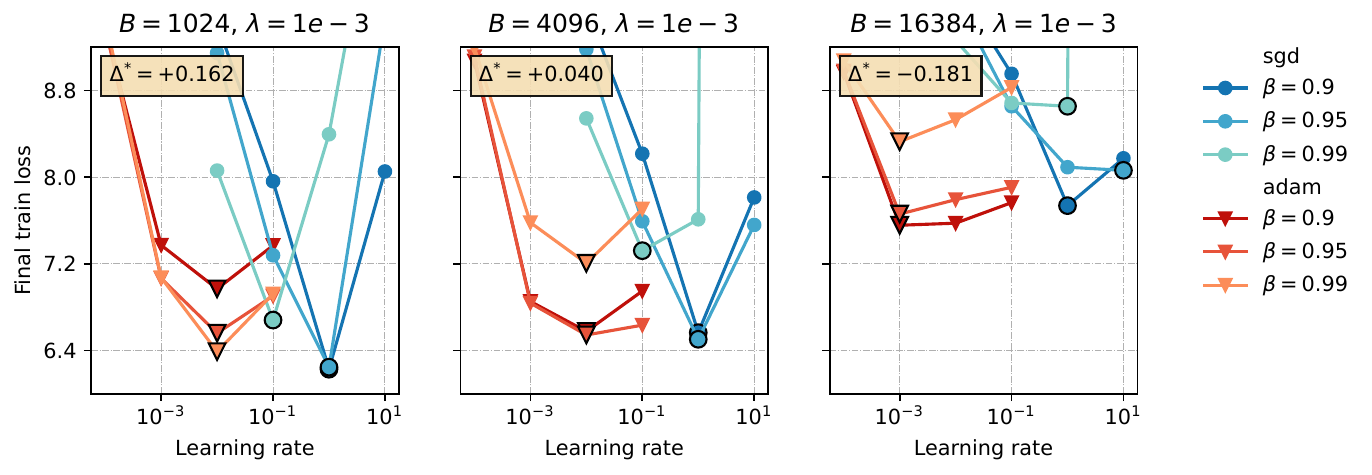}
  \includegraphics[width=0.6\linewidth]{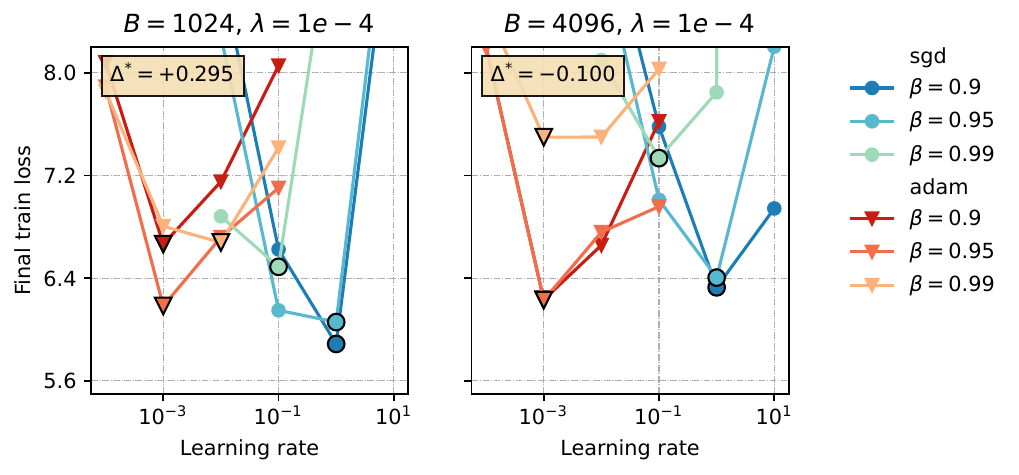}
  \caption{ViT model trained on I21K with \emph{independent} weight decay $\lambda$~\citep{wortsman2024small}. We observe the same trend as~\Cref{fig:main}, that is the optimizer advantage shifts from SGD $\to$ Adam as the batch size scales.
  }
  \label{fig:i21k_vit_wd}
\end{figure}

\paragraph{GPT on FineWeb.} Similarly to~\Cref{app:exp_fineweb}, we use a GPT model (see~\Cref{app:arch}) trained on FineWeb 1B tokens. We use \texttt{bfloat16} precision, 10\% linear warmup with cosine scheduler to 0, and global norm clipping to 1 for both Adam and SGD. We use \emph{independent} weight decay~\citep{wortsman2024small}, which is decoupled and independent of learning rate $\eta$.

We tune the learning rate $\eta$, momentum $\beta$, and batch size $B$, also adding the weight decay $\lambda$ as
\begin{itemize}[leftmargin=*]
    \item SGD (32 runs)
        \begin{align*}
        (\eta, \beta, B, \lambda) &\in \{10^{-2}, 10^{-1}, 10^{0}, 10^{1}\}\\
                      &\times \{0.9, 0.95, 0.99, 0.999\}\\
                      &\times \{16, 1024\}\\
                      &\times \{10^{-4}\}.
        \end{align*}
    \item Adam (32 runs) ($\beta_1=\beta_2=\beta$) 
        \begin{align*}
        (\eta, \beta, B, \lambda) &\in \{10^{-4}, 10^{-3}, 10^{-2}, 10^{-1}\}\\
                      &\times \{0.9, 0.95, 0.99, 0.999\}\\
                      &\times \{16, 1024\}\\
                      &\times \{10^{-4}\}.
        \end{align*}
\end{itemize}
This setup trains a total of 64 model configurations. 

\begin{figure}[!ht]
  \centering
  \includegraphics[width=0.6\linewidth]{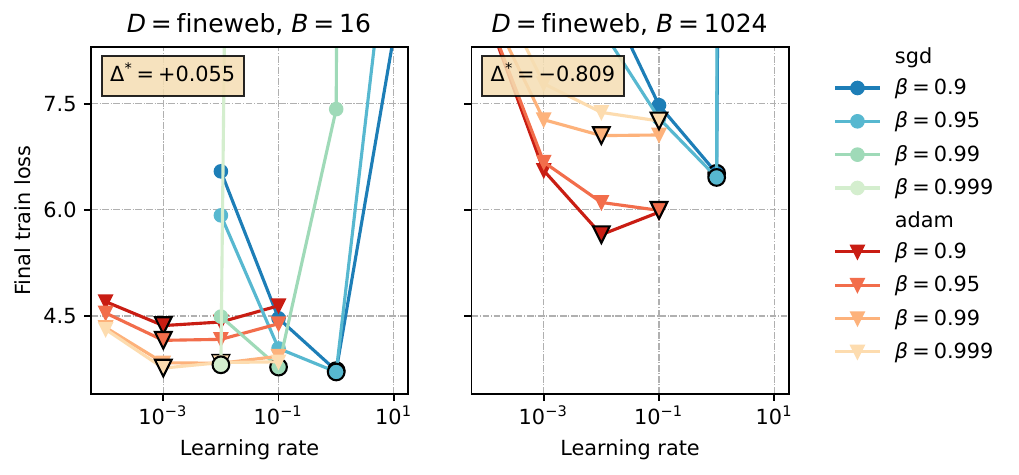}
  \caption{GPT model trained on FineWeb with \emph{independent} weight decay $\lambda=1e-4$~\citep{wortsman2024small}. We observe the same trend as~\Cref{fig:main}, that is the optimizer advantage shifts from SGD to Adam as the batch size scales.
  }
  \label{fig:fineweb_wd}
\end{figure}

\newpage
\subsection{Effect of number of steps}\label{app:exp_steps}

\begin{figure}[h]
  \centering
  \includegraphics[width=0.3\linewidth]{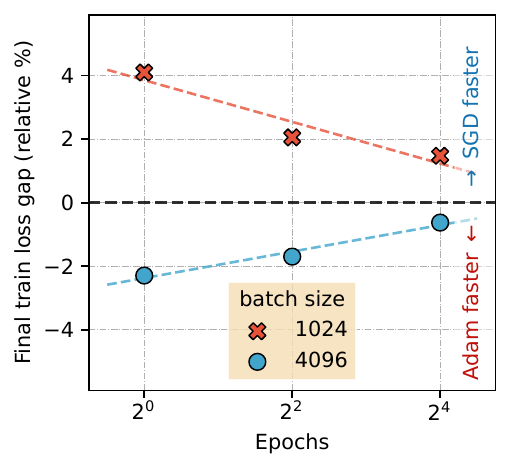}
  \caption{ViT trained on I21K for multiple epochs under a fixed $B$. More iterations reduces $\Delta$.}
  \label{fig:gap_i21k_vit_steps}
\end{figure}

Throughout this paper, we presented the batch size sweep results~(\Cref{fig:main}) that fix the total number of training samples (or total token budget). Here, we conduct a preliminary study of yet another axis: under fixed batch size $B$, how does the Adam--SGD gap change when scaling the number of steps? We train ViT on I21K for multiple epochs.~\Cref{fig:gap_i21k_vit_steps} shows that the gap decreases towards $\Delta=0$ regardless of whether the initial advantage is for SGD or Adam, suggesting that more iterations reduce the gap. 

For the experiments in~\Cref{fig:gap_i21k_vit_steps}, we use the same ViT model from~\Cref{app:arch} but trained for a longer horizon. In particular, we use \texttt{bfloat16} precision, 10\% linear warmup with cosine scheduler to 0, and global norm clipping to 1 for both Adam and SGD. We train the model for 4 and 16 epochs.

\textbf{Epochs 4.} We tune the learning rate $\eta$, momentum $\beta$, and batch size $B$ as
\begin{itemize}[leftmargin=*]
    \item SGD (32 runs)
        \begin{align*}
        (\eta, \beta, B) &\in \{10^{-2}, 10^{-1}, 10^{0}, 10^{1}\}\\
                      &\times \{0.9, 0.95, 0.99, 0.999\}\\
                      &\times \{1024, 4096\}.
        \end{align*}
    \item Adam (32 runs) ($\beta_1=\beta_2=\beta$) 
        \begin{align*}
        (\eta, \beta, B) &\in \{10^{-4}, 10^{-3}, 10^{-2}, 10^{-1}\}\\
                      &\times \{0.9, 0.95, 0.99, 0.999\}\\
                      &\times \{1024, 4096\}.
        \end{align*}
\end{itemize}
\textbf{Epochs 16.} We tune the learning rate $\eta$, momentum $\beta$, and batch size $B$ as
\begin{itemize}[leftmargin=*]
    \item SGD (18 runs)
        \begin{align*}
        (\eta, \beta, B) &\in \{10^{-2}, 10^{-1}, 10^{0}\}\\
                      &\times \{0.9, 0.95, 0.99\}\\
                      &\times \{1024, 4096\}.
        \end{align*}
    \item Adam (18 runs) ($\beta_1=\beta_2=\beta$) 
        \begin{align*}
        (\eta, \beta, B) &\in \{10^{-4}, 10^{-3}, 10^{-2}\}\\
                      &\times \{0.9, 0.95, 0.99\}\\
                      &\times \{1024, 4096\}.
        \end{align*}
\end{itemize}
For 16 epochs, training one run takes $\approx60$h on an GPU A100-80GB. Therefore, we excluded the consistently worst-performing settings from epochs 1 and 4: $\beta=0.999$ and $\eta$ values. 

This setup trains a total of 100 model configurations.

\textbf{Epochs 4 (ResNet50).} Lastly, we train ResNet50 also for a larger number of steps. We tune the learning rate $\eta$, momentum $\beta$, and batch size $B$ as
\begin{itemize}[leftmargin=*]
    \item SGD (60 runs)
        \begin{align*}
        (\eta, \beta, B) &\in \{10^{-2}, 10^{-1}, 10^{0}, 10^1, 3\cdot10^1\}\\
                      &\times \{0.9, 0.95, 0.99, 0.999\}\\
                      &\times \{4096, 16384, 65536\}.
        \end{align*}
    \item Adam (60 runs) ($\beta_1=\beta_2=\beta$) 
        \begin{align*}
        (\eta, \beta, B) &\in \{10^{-4}, 10^{-3}, 10^{-2}, 3\cdot10^{-2}, 10^{-1}\}\\
                      &\times \{0.9, 0.95, 0.99, 0.999\}\\
                      &\times \{4096, 16384, 65536\}.
        \end{align*}
\end{itemize}

\begin{figure}[h]
  \centering
  \includegraphics[width=0.8\linewidth]{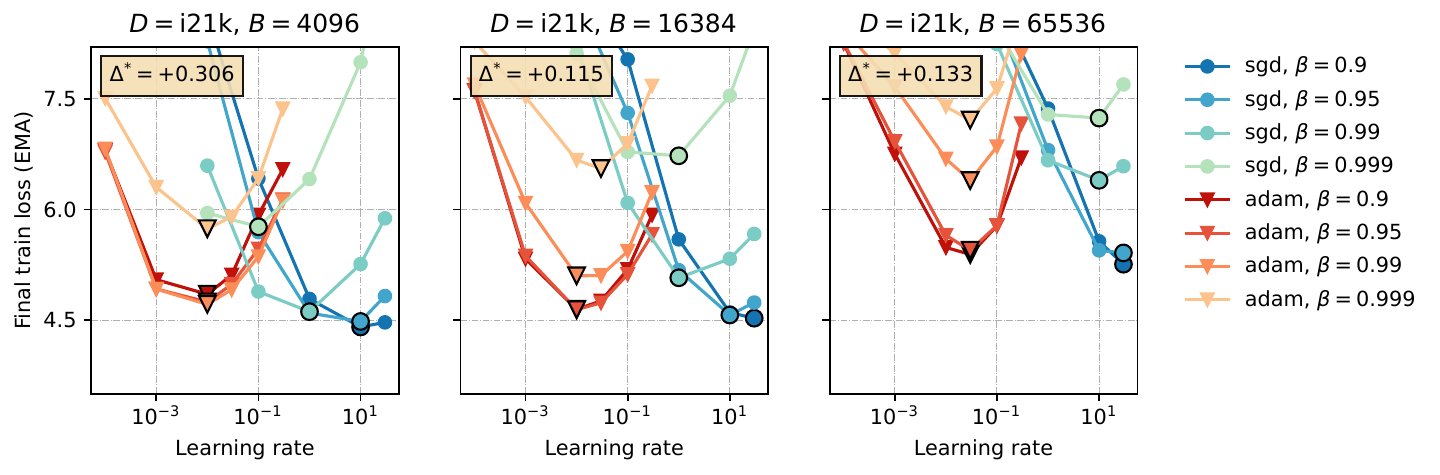}
  \caption{ResNet50 trained on I21K for 4 epochs  across $B$. Note that even at a large $B=65536$, SGD still has the advantage, suggesting a gap floor that depends on the architecture an data.}
  \label{fig:gap_i21k_resnet50_steps}
\end{figure}

\newpage
\subsection{Effect of independent \texorpdfstring{$\beta$}{beta}s}\label{app:independent_betas}

To ablate our prior results that rely on $\beta=\beta_1=\beta_2$~\citep{orvieto2025search}, we run independent $\beta$s sweep at small batch size. We use the same setup as~\Cref{app:exp_fineweb}.

We tune the learning rate $\eta$, $\beta_1$, $\beta_2$, and batch size $B$ as

\begin{itemize}[leftmargin=*]
    \item Adam (240 runs) ($\beta_1 \neq \beta_2$) 
        \begin{align*}
        (\eta, \beta_1, \beta_2,B) &\in \{3\cdot10^{-5}, 10^{-4},3\cdot10^{-4}, 10^{-3}, 3\cdot10^{-3}\}\\
                      &\times \{0.9, 0.95, 0.99, 0.999\}\\
                      &\times \{0.9, 0.95, 0.99, 0.999, 0.9999, 0.99999\}\\
                      &\times \{1, 16\}.
        \end{align*}
\end{itemize}

\begin{figure}[!ht]
  \centering
  \includegraphics[width=\linewidth]{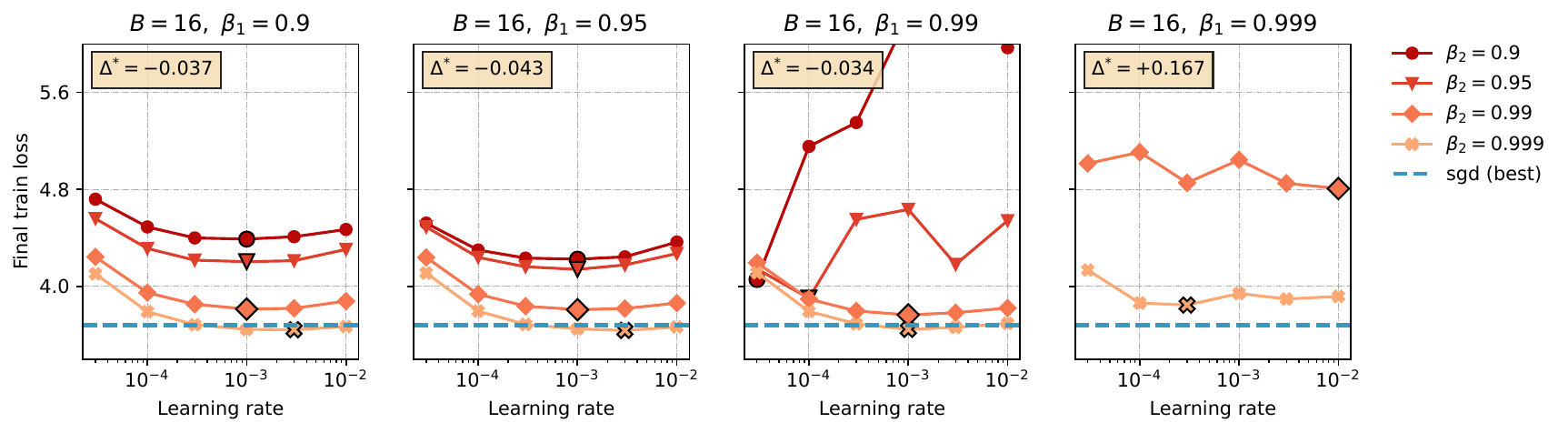}
  \includegraphics[width=\linewidth]{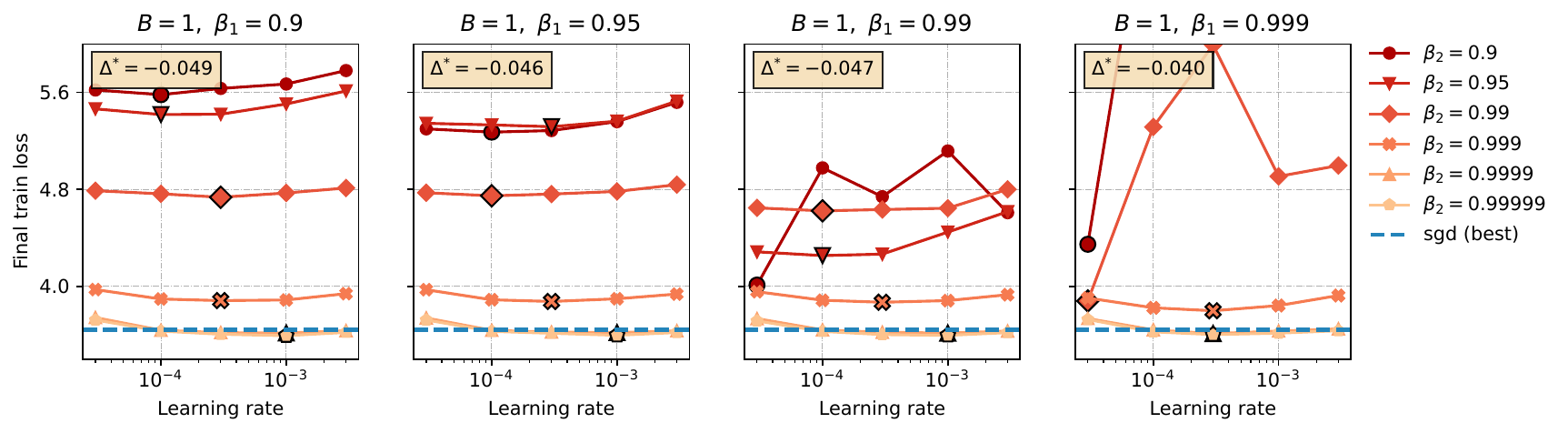}
  \caption{GPT trained on FineWeb with independent $\beta$s at $B=1$ and $B=16$.}
  \label{fig:fineweb_betas}
\end{figure}

\Cref{fig:fineweb_betas} shows that at $B=16$, a large $\beta_2=0.999$ leads to similar performance as SGD. The same is observed for $B=1$ but requiring an even larger $\beta_2=0.9999$. At first glance, this seems in contrast with prior results showing that SGD outperforms Adam at small $B$ for this setup. Note however, that Adam with a $\beta_2 \to 1$ corresponds essentially to SGD. 
Since SGD is optimal under small $B=1$, Adam with $\beta_2 \approx 1$ approximates SGD, suppressing its adaptivity. 

\newpage
\subsection{Large scale GPT and ViT}\label{app:scale}

We extend results in the main paper with larger scale architectures, showing that the \emph{crossover} batch size is observed even for larger scale GPT and ViT architectures. 

\textbf{GPT 250M on FineWeb 5BT.} Same setup as~\Cref{app:exp_fineweb} but larger scale. The GPT here has 24 layers, 12 heads, 768 embed size, and 1024 seq length. We tune the learning rate $\eta$, $\beta$, and batch size $B$ as

\begin{itemize}[leftmargin=*]
    \item SGD (64 runs)
        \begin{align*}
        (\eta, \beta, B) &\in \{10^{-2}, 10^{-1}, 10^{0}, 10^{1}\}\\
                      &\times \{0.9, 0.95, 0.99, 0.999\}\\
                      &\times \{16, 64, 256, 1024\}.
        \end{align*}
    \item Adam (64 runs) ($\beta_1=\beta_2=\beta$) 
        \begin{align*}
        (\eta, \beta, B) &\in \{10^{-4}, 10^{-3}, 10^{-2}, 10^{-1}\}\\
                      &\times \{0.9, 0.95, 0.99, 0.999\}\\
                      &\times \{16, 64, 256, 1024\}.
        \end{align*}
\end{itemize}

\vspace{-1em}
\begin{figure}[!ht]
  \centering
  \includegraphics[width=\linewidth]{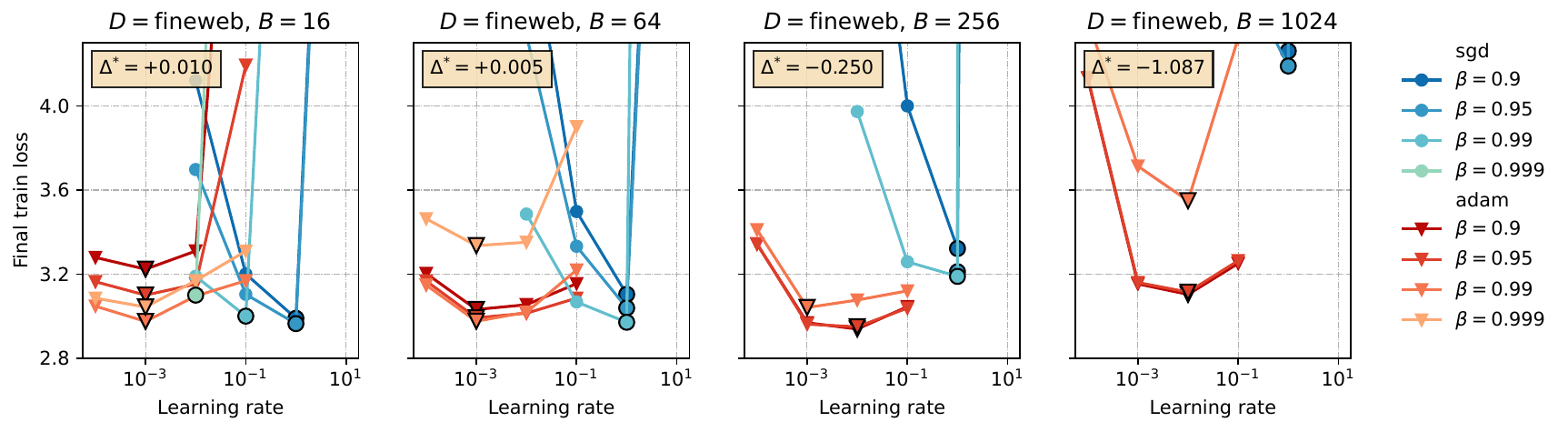}
  \caption{GPT 250M trained on FineWeb 5BT across batch sizes.}
  \label{fig:fineweb_large}
\end{figure}

\textbf{ViT 128M on I21K.} Same setup as~\Cref{app:exp_i21k} but larger scale. The ViT architecture here has 16 layers, 16 heads, and 768 embedding size. We tune the learning rate $\eta$, $\beta$, and batch size $B$ as

\begin{itemize}[leftmargin=*]
    \item SGD (32 runs)
        \begin{align*}
        (\eta, \beta, B) &\in \{10^{-2}, 10^{-1}, 10^{0}, 10^{1}\}\\
                      &\times \{0.9, 0.95, 0.99, 0.999\}\\
                      &\times \{1024, 4096\}.
        \end{align*}
    \item Adam (32 runs) ($\beta_1=\beta_2=\beta$) 
        \begin{align*}
        (\eta, \beta, B) &\in \{10^{-4}, 10^{-3}, 10^{-2}, 10^{-1}\}\\
                      &\times \{0.9, 0.95, 0.99, 0.999\}\\
                      &\times \{1024, 4096\}.
        \end{align*}
\end{itemize}

\vspace{-1em}
\begin{figure}[!ht]
  \centering
  \includegraphics[width=0.6\linewidth]{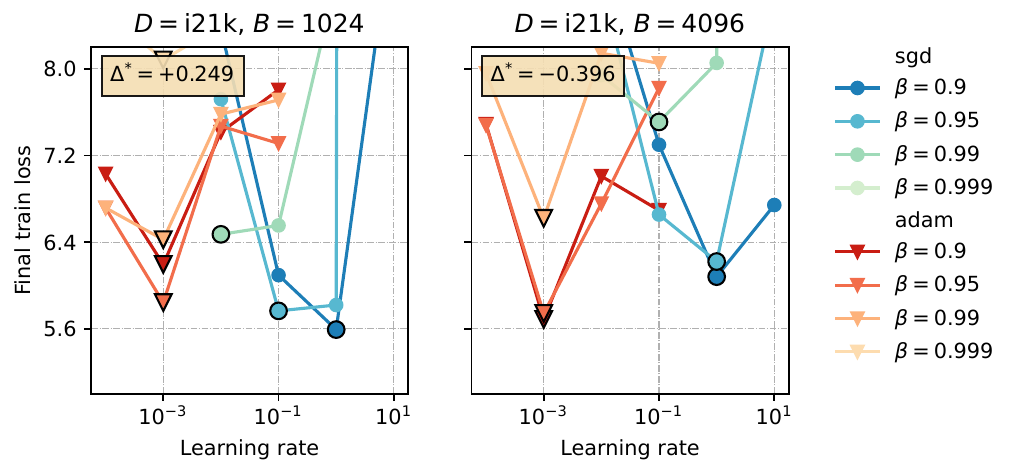}
  \caption{ViT 128M trained on I21K across batch sizes.}
  \label{fig:i21k_large}
\end{figure}

This setup trains a total of 192 model configurations.

\newpage
\subsection{Pushing the batch size to 1}\label{app:bs1}

Extending the~\Cref{app:exp_hg38}, we push the GPT trained on HG38 dataset using \texttt{char} tokenizer to batch size 1 and 4. We observe that even at the smallest batch size, the gap remains in favor of Adam. This suggests that, in certain setups, Adam may always be better.  

We tune the learning rate $\eta$, momentum $\beta$, and batch size $B$ as

\begin{itemize}[leftmargin=*]
    \item SGD (32 runs)
        \begin{align*}
        (\eta, \beta, B) &\in \{10^{-2}, 3\cdot10^{-2}, 10^{-1}, 3\cdot10^{-1}, 10^0\}\\
                      &\times \{0.9, 0.95, 0.99, 0.999\}\\
                      &\times \{1, 4\}.
        \end{align*}
    \item Adam (32 runs) ($\beta_1=\beta_2=\beta$) 
        \begin{align*}
        (\eta, \beta, B) &\in \{10^{-4}, 3\cdot10^{-4}, 10^{-3}, 3\cdot10^{-3}\}\\
                      &\times \{0.9, 0.95, 0.99, 0.999\}\\
                      &\times \{1, 4\}.
        \end{align*}
\end{itemize}

This setup trains a total of 64 model configurations.

\begin{figure}[!ht]
  \centering
  \includegraphics[width=0.6\linewidth]{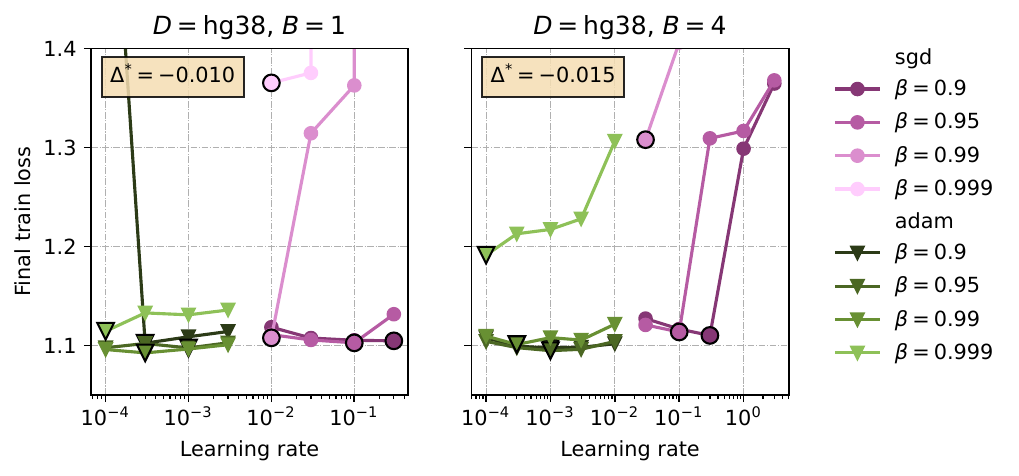}
  \caption{GPT trained on HG38 at smallest batch size still presents Adam$>$SGD.}
  \label{fig:hg38_bs1_train_loss}
\end{figure}

\newpage
\subsection{Validation performance gap}\label{app:exp_validation}
Throughout this work, we have focused on training loss. Here, we report the gap in generalization performance. These are the same models from~\Cref{fig:main}, that is GPT from~\Cref{app:exp_fineweb} and ViT from~\Cref{app:exp_i21k}.

\begin{figure}[!ht]
  \centering
  \includegraphics[width=\linewidth]{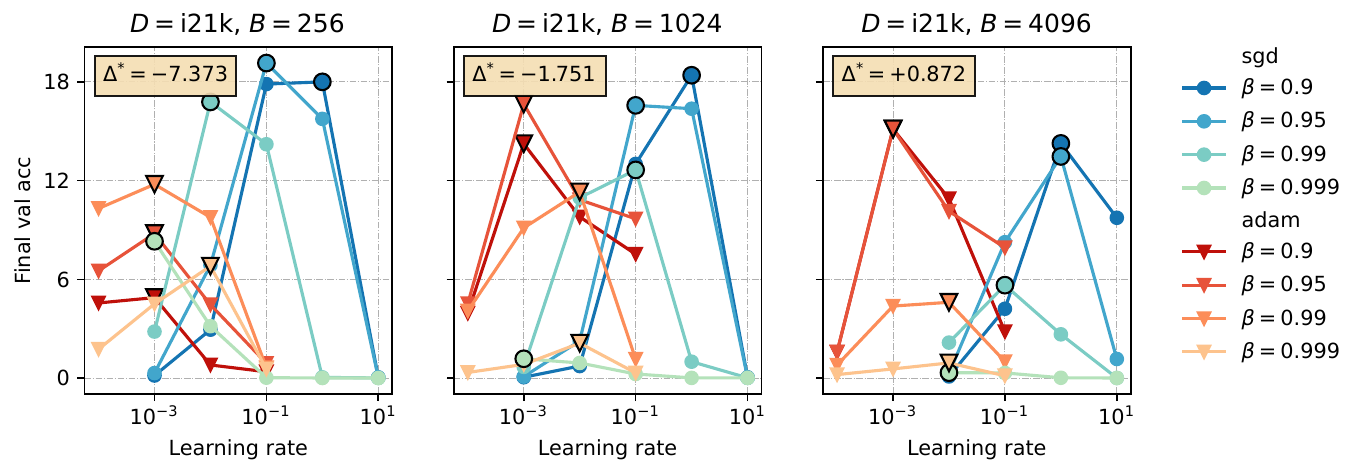}
  \caption{Top-1 validation accuracy of ViT model trained on I21K. We observe the same trend as~\Cref{fig:main}, that is the optimizer advantage shifts from SGD $\to$ Adam as the batch size scales.
  }
  \label{fig:validation_i21k}
\end{figure}
\begin{figure}[!ht]
  \centering
  \includegraphics[width=\linewidth]{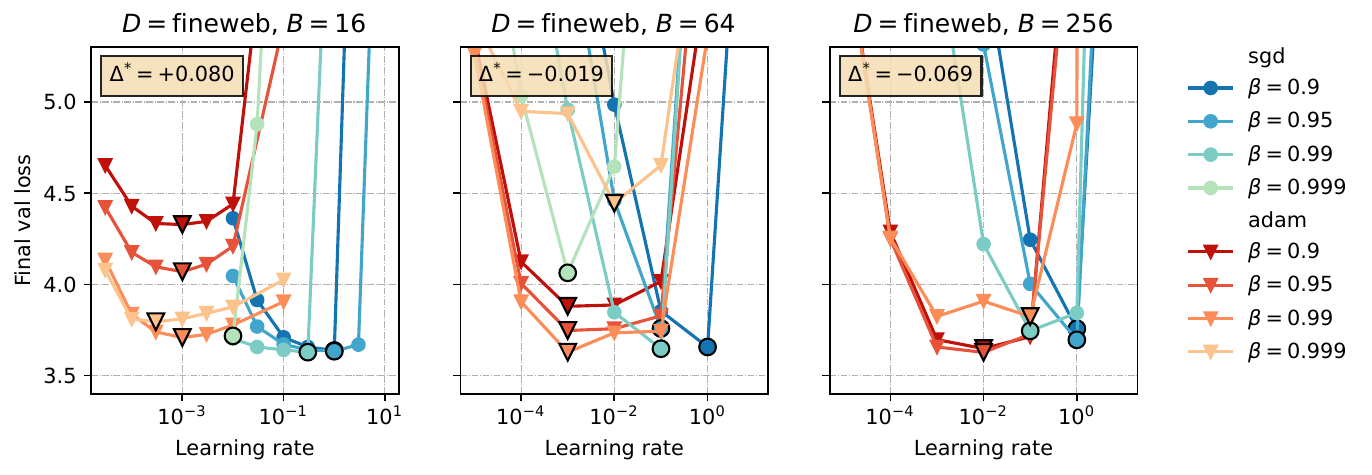}
  \caption{Validation loss of GPT model of 50M parameters trained on FineWeb 1B tokens. We observe the same trend as~\Cref{fig:main}, that is the optimizer advantage shifts from SGD $\to$ Adam as the batch size scales.
  }
  \label{fig:validation_fineweb}
\end{figure}

\newpage
\subsection{Architecture interpolation with independent components}
\label{app:exp_arch_interpolation_independent}
Same experimental settings as~\Cref{app:exp_arch_interpolation}, including the learning rate $\eta$ and momentum $\beta$ tuning grid.

We tune the learning rate $\eta$, momentum $\beta$, and batch size $B$ as
\begin{itemize}[leftmargin=*]
    \item SGD (12 runs)
        \begin{align*}
        (\eta, \beta, B) &\in \{10^{-2}, 10^{-1}, 10^0\}\\
                      &\times \{0.9, 0.95, 0.99, 0.999\}\\
                      &\times \{1024\}.
        \end{align*}
    \item Adam (12 runs) ($\beta_1=\beta_2=\beta$): 
        \begin{align*}
        (\eta, \beta, B) &\in \{10^{-4}, 10^{-3}, 10^{-2}\}\\
                      &\times \{0.9, 0.95, 0.99, 0.999\}\\
                      &\times \{1024\}.
        \end{align*}
\end{itemize}

\input{tables/arch_interpolation_independent}

This setup trains a total of 72 model configurations, as there are 3 configurations in~\Cref{tab:arch_interpolation_independent}.

\newpage
\subsection{Isolating early training phase with \texttt{stable\_sgd} shared trajectory}
\label{app:exp_early_training_sgd}
Same setup as in~\Cref{app:exp_early_training} but training the shared trajectory \texttt{stable\_sgd} instead of \texttt{stable\_adam}. This shared SGD trajectory uses $\eta=3\cdot10^{-1}$ and $\beta=0.95$. 

We tune the learning rate $\eta$, momentum $\beta$, and batch size $B$ at every branching as
\begin{itemize}[leftmargin=*]
    \item SGD (20 runs)
        \begin{align*}
        (\eta, \beta, B) &\in \{3\cdot10^{-2}, 10^{-1}, 3\cdot10^{-1}, 10^{0}, 3\cdot10^{0}\}\\
                      &\times \{0.9, 0.95, 0.99, 0.999\}\\
                      &\times \{1024\}.
        \end{align*}
    \item Adam (20 runs) ($\beta_1=\beta_2=\beta$): 
        \begin{align*}
        (\eta, \beta, B) &\in \{10^{-3}, 3\cdot10^{-3}, 10^{-2}, 3\cdot10^{-2}, 10^{-1}\}\\
                      &\times \{0.9, 0.95, 0.99, 0.999\}\\
                      &\times \{1024\}.
        \end{align*}
\end{itemize}

This setup trains a total of 280 model configurations, as there are 7 branches.

\begin{figure}[ht!]
  \centering
  \begin{subfigure}[c]{.40\linewidth}
    \centering
    \includegraphics[width=\linewidth]{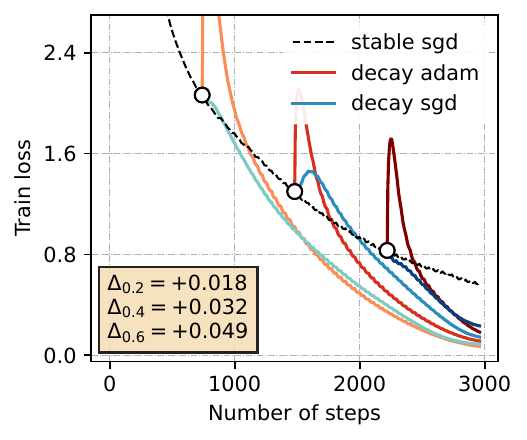}
    \caption{Overparametrized w/ TinyStories}
  \end{subfigure}%
  \begin{subfigure}[c]{.40\linewidth}
    \centering
    \includegraphics[width=\linewidth]{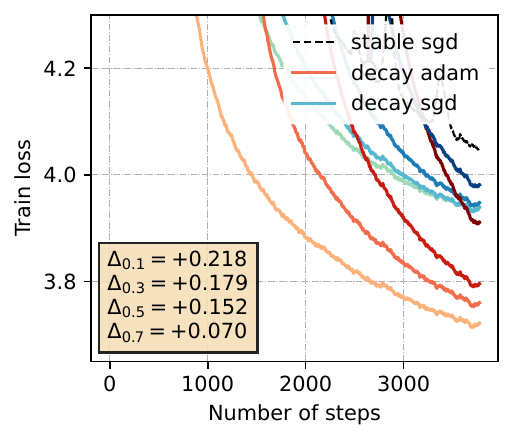}
    \caption{Underparametrized w/ FineWeb}
  \end{subfigure}
  \caption{GPT model of 50M parameters trained on (a) 10M TinyStories tokens for 80 epochs and (b) 1B FineWeb tokens for 1 epoch at batch size $B=1024$. 
  We isolate the effect of early training dynamics by branching Adam and SGD from a shared constant learning rate trajectory (\texttt{stable\_sgd}). 
  Branching happens at different intervals to probe different stages of training. (a) For overparametrization, both branches converge to low loss as the problem is solvable with more epochs, while (b) for underparametrization, SGD lags behind.
  For every branch, the momentum state is reset, and the learning rate $\eta$ and momentum $\beta$ are tuned, see~\Cref{app:exp_early_training}.
  }
  \label{fig:early_training_sgd}
\end{figure}

%%%%%%%%%%%%%%%%%%%%%%%%%%%%%%%%%%%%%%%%%%%%%%%%%%%%%%%%%%%%

\newpage
\section{Detailed theoretical study}\label{app:theory}

To provide theoretical evidence on why the Adam--SGD gap decreases with batch size and the potential existence of a crossover batch size, we use the non-convex convergence bounds from \citep{kovalev2025understanding}, i.e., the bounds on the gradient norm. These bounds were found to have a strong predictive ability for the scaling of the algorithm's parameters with the token budget $T$ \citep{shulgin2026deriving,islamov2026role}.

\subsection{Notation and setting}

Our goal is to solve the stochastic optimization problem
\begin{equation}\label{eq:problem_app}
    \min_{x\in\gX} \left[f(x) = \E_{\xi\sim\gD}[f(x,\xi)]\right],
\end{equation}
where the space $\gX$ is equipped with an inner product $\left<\cdot,\cdot\right>$ and a norm $\|\cdot\|$, which need not be the norm induced by this inner product. To solve \Cref{eq:problem_app}, we use the Stochastic Trust-Region Gradient Method, summarized in \Cref{alg}. Depending on the choice of $\|\cdot\|$, this framework recovers several practical optimization algorithms. For example, when $\|\cdot\|=\|\cdot\|_2$, the update in \Cref{alg} reduces to
\[
x_{k+1} = x_k -\eta\frac{m_{k+1}}
{\|m_{k+1}\|_2},
\] 
which corresponds to a step of normalized SGD (NSGD) with momentum. When $\|\cdot\|=\|\cdot\|_\infty$, the update recovers SignSGD with momentum,
\[
x_{k+1} = x_k - \eta\mathrm{sign}(m_{k+1}).
\]
In our experiments, we approximate the convergence behavior of SGD with momentum and clipping by that of normalized SGD with momentum. Similarly, we use SignSGD with momentum as a proxy for Adam, motivated by their empirical similarity~\citep{orvieto2025search}.

\subsection{Convergence bounds}

\begin{algorithm}[b]
  \caption{Stochastic Trust-region Gradient Method with Momentum}
  \label{alg}
  \begin{algorithmic}
    \STATE {\bfseries Input:} norm $\|\cdot\|$, step size $\eta > 0$, momentum $\alpha \in (0,1)$, number of iterations $K \in \{1,2,\dots\}$
    \FOR{$k=1, \dots, K-1$}
    \STATE Sample $\xi_k\in\gD$
    \STATE Compute $m_{k+1} = (1-\alpha)m_k + \alpha \nabla f(x_k, \xi_k)$
    \STATE Compute $d_{k+1} = \mathrm{argmin}_d \left<d, m_{k+1}\right>$ such that $\|d\|\le 1$
    \STATE Update $x_{k+1} = x_k - \eta d_{k+1}$
    \ENDFOR
    \STATE return
  \end{algorithmic}
\end{algorithm}

Assuming that 
\begin{itemize}
    \item $\|\nabla f(x) - \nabla f(y)\|_* \le L\|x-y\|$ for all $x,y\in\R^d$,
    \item there exists $\rho > 0$ such that $\|x\|_* \le\rho\|x\|_2$, and 
    \item there exists $\sigma^2$ such that $\E_{\xi\sim\gD}[\|\nabla f(x,\xi)-\nabla f(x)\|^2_2] \le \frac{\sigma^2}{B}$, where $B$ is the batch size,
\end{itemize}
\citet{kovalev2025understanding} provide the following non-convex bounds
\begin{align}
    \min_{1\le k \le K}\E[\|\nabla f(x_k)\|_*] &\lesssim \frac{\delta_0}{\eta K} + \frac{\rho\sigma}{\alpha \sqrt{B}K} + \frac{\sqrt{\alpha}\rho\sigma}{\sqrt{B}} + L\eta + \frac{L\eta}{\alpha}\notag\\
    &= \frac{\delta_0 B}{\eta T} + \frac{\rho\sigma\sqrt{B}}{\alpha T} + \frac{\sqrt{\alpha}\rho\sigma}{\sqrt{B}} + L\eta + \frac{L\eta}{\alpha},\label{eq:bound}
\end{align}
where $K$ is the number of optimization steps, $T$ is the total budget, $\delta_0 = f(x_0) - f^*$, and numerical constants are ignored. We use the connection $T=KB$ to obtain the second line in~\Cref{eq:bound}. Following derivations of \citet{shulgin2026deriving} in the large $T\gg1$ and small $\alpha$ regime, we obtain that the optimal $\eta$ and $\alpha$ are 
\begin{equation}
    \eta \propto \frac{B\delta_0^{3/4}}{L^{1/4}(\rho\sigma)^{1/2}T^{3/4}}, \quad \alpha \propto \frac{\sqrt{L\delta_0}B}{\rho\sigma\sqrt{T}}.
\end{equation}
Using this choice of hyperparameters, we obtain a bound minimized with respect to both $\eta$ and $\alpha$
\begin{align}\label{eq:minimized_bound}
    U_{\|\cdot\|} = \min_{1\le k \le K}\E[\|\nabla f(x_k)\|_*] &\lesssim \frac{(L\delta_0)^{1/4}\sqrt{\rho\sigma}}{T^{1/4}} 
    + \frac{(\rho\sigma)^2}{\sqrt{L\delta_0TB}}
    + \frac{(L\delta_0)^{3/4}B}{\sqrt{\rho\sigma}T^{3/4}}.
\end{align}

\subsection{Comparison of the bounds for NSGD and SignSGD}

Now we compare the bound \eqref{eq:minimized_bound} for NSGD and SignSGD with momentum using the specific choice of constants. For the $\ell_2$ norm, we have $L=L_2$ and $\rho=1$, while for the $\ell_\infty$, it is $L=L_\infty$ and $\rho=\sqrt{d}$. Since we solve the same problem, the noise $\sigma$ is the same for both methods. Therefore, the difference in the bounds for NSGD and SignSGD with momentum is\footnote{We denote by $U_2$ and $U_\infty$ the bounds obtained from \eqref{eq:minimized_bound} with the norm chosen as $\ell_2$ and $\ell_\infty$.}
% \begin{align}\label{eq:gap_app}
%     U_2 - U_\infty &= \frac{(\sigma^2\delta_0)^{1/4}(L_2^{1/4} - (L_\infty d)^{1/4})}{T^{1/4}}
%     + \frac{\sigma^2}{\sqrt{\delta_0TB}}\left(\frac{1}{\sqrt{L_2}} - \frac{d}{\sqrt{L_\infty}}\right)\notag\\
%     &\hspace{7cm}+ \frac{\delta_0^{3/4}B}{\sqrt{\sigma}T^{3/4}}\left(L_2^{3/4} - \frac{L_\infty^{3/4}}{d^{1/4}}\right).
% \end{align}
\begin{equation}\label{eq:gap_app}
\begin{aligned}
\Delta_U
&= U_\infty-U_2 \\
&= \frac{(\sigma^2\delta_0)^{1/4}\left((L_\infty d)^{1/4}-L_2^{1/4}\right)}{T^{1/4}}
+\frac{\sigma^2}{\sqrt{\delta_0TB}}
\left(\frac{d}{\sqrt{L_\infty}}-\frac{1}{\sqrt{L_2}}\right) +\frac{\delta_0^{3/4}B}{\sqrt{\sigma}T^{3/4}}
\left(\frac{L_\infty^{3/4}}{d^{1/4}}-L_2^{3/4}\right).
\end{aligned}
\end{equation}
The derivative of the gap with respect to the batch size is given below
% \begin{align}
%     \frac{\partial (U_2-U_\infty)}{\partial B} = - \frac{\sigma^2}{\sqrt{\delta_0TB^3}}\left(\frac{1}{\sqrt{L_2}} - \frac{d}{\sqrt{L_\infty}}\right) + \frac{\delta_0^{3/4}}{\sqrt{\sigma}T^{3/4}}\left(L_2^{3/4} - \frac{L_\infty^{3/4}}{d^{1/4}}\right).
% \end{align}
\begin{align}
    \frac{\partial \Delta_U}{\partial B} = -\frac{\sigma^2}{\sqrt{\delta_0TB^3}}\left(\frac{d}{\sqrt{L_\infty}} - \frac{1}{\sqrt{L_2}}\right) + \frac{\delta_0^{3/4}}{\sqrt{\sigma}T^{3/4}}\left(\frac{L_\infty^{3/4}}{d^{1/4}} - L_2^{3/4}\right).
\end{align}
We observe that the gap between the two optimizers and how it changes with batch size depends on the interaction of the geometry and dimensionality of the problem. Different relations between $L_2,L_\infty,$ and $d$ can lead to different signs of the gap, as well as different dynamics with the batch size. Let us consider several cases. Note that we always have $L_2 \le L_{\infty} \le d L_2$.

\paragraph{$\mathbf{1} < \mathbf{L}_{\boldsymbol{\infty}}/\mathbf{L_2} < \mathbf{d^{1/3}}$.}

In this scenario, we have 
\begin{align}\label{eq:second_case}
\Delta_U \propto \frac{1}{T^{1/4}} + \frac{1}{\sqrt{TB}} - \frac{B}{T^{3/4}},
\quad
\frac{\partial \Delta_U}{\partial B} \propto -\frac{1}{\sqrt{TB^3}} - \frac{1}{T^{3/4}}.
\end{align}
We observe that the derivative of the gap is always negative. Therefore, the sign of the gap also depends on the value of the gap at $B=1$. If the gap at the batch size $B=1$ is already negative, SignSGD is already favored at $B=1$ and remains favored as $B$ increases. If the gap at $B=1$ is positive, then we will observe that there is a crossover from NSGD being better to SignSGD being better. Note that the value of the gap at $B=1$ depends on the relation between $L_2, L_\infty,d$, as well as $\sigma$ and $\delta_0$. This serves as an additional support that the model, dataset, and gradient noise jointly affect the sign of the gap.

We observe that most of the experimental setups tested in our work lie in this regime, since we observe a crossover from SGD being better to Adam being better. Moreover, for the Genomics setting, we do observe that Adam is better than SGD even for a batch size $1$, which is also possible according to the convergence analysis.

\paragraph{$\mathbf{d^{1/3}} < \mathbf{L}_{\boldsymbol{\infty}}/\mathbf{L_2} < \mathbf{d}$.}

In this regime, we have 
\begin{align}
\Delta_U \propto \frac{1}{T^{1/4}} + \frac{1}{\sqrt{TB}} + \frac{B}{T^{3/4}},
\quad
\frac{\partial \Delta_U}{\partial B} \propto -\frac{1}{\sqrt{TB^3}} + \frac{1}{T^{3/4}}.
\end{align}

The relation between $L_2, L_\infty,$ and $d$ implies that the gap is always positive, i.e., NSGD achieves lower loss than SignSGD, and there exists a batch size $B \propto T^{1/6}$ when the gap is the smallest. This case might be linked to ResNet50/ImageNet-21K experiments, where we observe that SGD remains better than Adam even at a large batch size.

% \newpage
% \input{checklist.tex}

\end{document}

%% file: math_commands.tex
\usepackage{amsmath,amsfonts,bm}

\def\eqref#1{equation~\ref{#1}}
\def\1{\bm{1}}

\DeclareMathAlphabet{\mathsfit}{\encodingdefault}{\sfdefault}{m}{sl}
\SetMathAlphabet{\mathsfit}{bold}{\encodingdefault}{\sfdefault}{bx}{n}

\def\gD{{\mathcal{D}}}

\def\gX{{\mathcal{X}}}

\newcommand{\E}{\mathbb{E}}

\newcommand{\R}{\mathbb{R}}

%% file: tables/arch_hybrid.tex
\begin{table}[b!]
\vspace{-8mm}
\centering
\begin{minipage}{0.6\textwidth}
    \caption{GPT 50M parameters trained on 1B FineWeb tokens with a hybrid optimizer: SGD used for all layers except those in ``Adam layers''. The $\Delta$ measures the difference between the best Adam-only run and each hybrid configuration ($\Delta<0$ means Adam has lower loss). Each setup tunes the learning rate and momentum at $B=1024$, see~\Cref{app:exp_hybrid}.\\}
    \label{tab:hybrid_optimizer}
    \small\centering
    \begin{tabular}{lrrr}
    \toprule
    \textbf{Adam layers} & \multicolumn{1}{c}{\textbf{Train loss}} & \multicolumn{1}{c}{$\boldsymbol{\Delta}$} & \multicolumn{1}{c}{$\boldsymbol{\Delta\%}$} \\
    \midrule
    - & $3.995$ & $-0.312$ & $-8.47\%$ \\
    N (Norm) & $3.962$ & $-0.279$ & $-7.57\%$ \\
    E (Embed) & $3.891$ & $-0.208$ & $-5.65\%$ \\
    H (Head) & $3.780$ & $-0.097$ & $-2.63\%$ \\
    N + H~\citep{zhao2025deconstructing} & $3.737$ & $-0.054$ & $-1.47\%$ \\
    N + H + E (ours) & $3.719$ & $-0.036$ & $-0.98\%$ \\
    All & $3.683$ & $0$ & $0\%$ \\
    \bottomrule
    \end{tabular}
\end{minipage}
\hfill
\begin{minipage}{0.37\textwidth}
    \centering
    \vspace{1em}
    \includegraphics[width=0.9\linewidth]{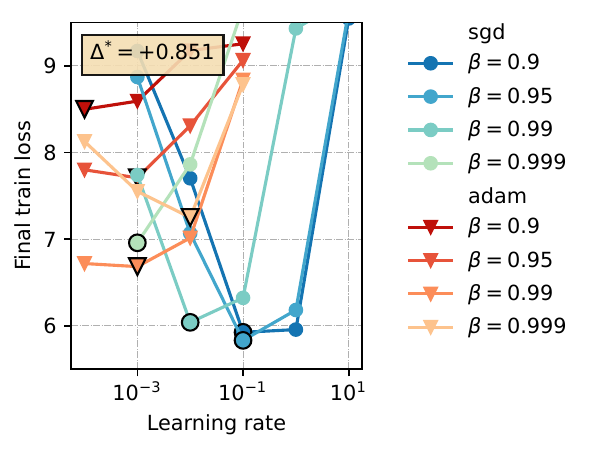}
    \captionof{figure}{ViT 18M parameters on I21K for one epoch at $B=256$. Despite component heterogeneity, SGD reaches $\approx12\%$ lower final loss than Adam, suggesting that component heterogeneity alone does not \emph{always} imply an Adam advantage.}
    \label{fig:gap_i21k_vit}
\end{minipage}
\end{table}

%% file: tables/arch_interpolation.tex
\definecolor{deltaA}{rgb}{0.0771,0.4549,0.6980}
\definecolor{deltaB}{rgb}{0.3412,0.7216,0.8157}
\definecolor{deltaC}{rgb}{0.9922,0.7660,0.5516}
\definecolor{deltaD}{rgb}{0.9882,0.5529,0.3490}
\definecolor{deltaE}{rgb}{0.7490,0.0627,0.0405}

\begin{wraptable}[19]{r}{0.53\textwidth}
% \vspace{-1.em}
\centering
\caption{Architecture interpolation from ResNet50 to ConvNext trained on I21K at $B=1024$. Three architecture choices (L: LayerNorm, G: GeLU, and D: Depthwise) are identified to shift the advantage from SGD to Adam, and are progressively stacked on top of one another. The $\Delta$ measures the difference between the best run trained with SGD and Adam ($\Delta<0$ means Adam has lower loss). Each setup tunes the learning rate and momentum, see~\Cref{app:exp_arch_interpolation}.}
\label{tab:arch_interpolation}
\small
\begin{tabular}{lrrrr}
    \toprule
    & \multicolumn{4}{c}{\textbf{Train loss}} \\
    \cmidrule(lr){2-5}
    \multicolumn{1}{c}{\textbf{Arch.}} & \multicolumn{1}{c}{\textbf{Adam}} & \multicolumn{1}{c}{\textbf{SGD}} & \multicolumn{1}{c}{$\boldsymbol{\Delta}$} & \multicolumn{1}{c}{$\boldsymbol{\Delta\%}$} \\
    \midrule
    ResNet50       & $6.133$ & $5.337$ & $+0.797$ & \textcolor{deltaA}{$+13.00\%$} \\
    + L    & $7.725$ & $7.162$ & $+0.563$ & \textcolor{deltaB}{$+7.29\%$} \\
    + L + G         & $6.275$ & $6.306$ & $-0.031$ & \textcolor{deltaC}{$-0.49\%$} \\
    + L + G + D     & $6.071$ & $6.791$ & $-0.720$ & \textcolor{deltaD}{$-11.86\%$} \\
    ConvNeXt       & $5.955$ & $6.880$ & $-0.925$ & \textcolor{deltaE}{$-15.53\%$} \\
    \bottomrule
\end{tabular}
\end{wraptable}

%% file: tables/arch.tex
\begin{table}[!ht]
\caption{Detailed overview of the main experimental setups proposed to study the Adam--SGD gap.}
\label{tab:arch}
\centering
\footnotesize
\begin{tabular}{lllc}
    \toprule
    \textbf{Architecture} & \textbf{Dataset} & \textbf{Modality} & \textbf{Transformer} \\
    \midrule
    GPT~\citep{radford2019language} & TinyStories~\citep{eldan2023tinystories} & Language & \cmark \\
    GPT~\citep{radford2019language} & Fineweb~\citep{penedo2024fineweb} & Language & \cmark \\
    GPT~\citep{radford2019language} & HG38~\citep{hg38} & Genomics & \cmark \\
    ViT~\citep{dosovitskiy2020image} & HT-I1K~\citep{kunstner2024heavy} & Vision & \cmark \\
    ViT~\citep{dosovitskiy2020image} & I21K~\citep{deng2009imagenet} & Vision & \cmark \\
    GRIT~\citep{ma2023grit} & ZINC250K~\citep{irwin2012zinc} & Graphs & \cmark \\
    \midrule
    GCNN~\citep{dauphin2017language} & TinyStories~\citep{eldan2023tinystories} & Language & \xmark \\
    GCNN~\citep{dauphin2017language} & Fineweb~\citep{penedo2024fineweb} & Language & \xmark \\
    GDN~\citep{yang2025gated} & HG38~\citep{hg38} & Genomics & \xmark \\
    ConvNext~\citep{liu2022convnet} & I21K~\citep{deng2009imagenet} & Vision & \xmark \\
    ResNet~\citep{he2016deep} & I21K~\citep{deng2009imagenet} & Vision & \xmark \\
    GAT~\citep{velivckovic2018graph} & ZINC250K~\citep{irwin2012zinc} & Graphs & \xmark \\
    \bottomrule
\end{tabular}
\end{table}

%% file: tables/arch_interpolation_independent.tex
\begin{table}[ht!]
\centering
\caption{Single architecture-choice ablation (L: LayerNorm, G: GeLU, and D: Depthwise) at $B=1024$. Each individual modification slightly moves the advantage towards Adam. However, as we have seen in~\Cref{tab:arch_interpolation}, the cumulative effect is even stronger. The $\Delta$ measures the difference between the best run trained with SGD and Adam ($\Delta<0$ means Adam has lower loss).\\}
\label{tab:arch_interpolation_independent}
\begin{tabular}{lrrrr}
    \toprule
    & \multicolumn{4}{c}{\textbf{Train loss}} \\
    \cmidrule(lr){2-5}
    \multicolumn{1}{c}{\textbf{Arch.}} & \multicolumn{1}{c}{\textbf{Adam}} & \multicolumn{1}{c}{\textbf{SGD}} & \multicolumn{1}{c}{$\boldsymbol{\Delta}$} & \multicolumn{1}{c}{$\boldsymbol{\Delta\%}$} \\
    \midrule
    ResNet50       & $6.133$ & $5.337$ & $+0.797$ & $+13.00\%$ \\
    + L            & $7.725$ & $7.162$ & $+0.563$ & $+7.29\%$ \\
    + G            & $5.817$ & $5.233$ & $+0.584$ & $+11.16\%$ \\
    + D            & $6.002$ & $5.348$ & $+0.654$ & $+12.23\%$ \\
    \bottomrule
\end{tabular}
\end{table}